\documentclass[5p, times]{elsarticle}
\makeatletter
\def\ps@pprintTitle{%
 \let\@oddhead\@empty
 \let\@evenhead\@empty
 \def\@oddfoot{Published in Pattern Recognition, \url{https://doi.org/10.1016/j.patcog.2025.111788}\hfill}%
 \let\@evenfoot\@oddfoot}
\makeatother

\usepackage{setspace}
\usepackage[utf8]{inputenc} 
\usepackage[table]{xcolor}
\definecolor{lightgray}{gray}{0.9}
\usepackage{subcaption}
\usepackage{balance}
\usepackage{amsmath,amssymb}
\usepackage{url}
\usepackage{xfrac}
\journal{Pattern Recognition}

\biboptions{sort&compress}

\usepackage{color}
\usepackage{xcolor}

\usepackage{xspace}
\usepackage{graphicx}
\usepackage{longtable}
\usepackage{caption}
\usepackage{amsmath,amsthm}
\usepackage{multirow}
\usepackage{multicol}
\usepackage{booktabs}
\usepackage{url}
\usepackage{pifont}
\usepackage{color}
\usepackage{supertabular}
\usepackage[export]{adjustbox}

\usepackage{algorithm}
\usepackage{algorithmicx}
\usepackage{algpseudocode}

\usepackage{amsmath}
\usepackage{amsthm,amsmath,amssymb}
\usepackage{bm}
\usepackage{subcaption}
\usepackage{hyperref}
\usepackage{CJKutf8}
\usepackage{booktabs}
\usepackage{amssymb}
\usepackage{pifont}
\newcommand{\xmark}{\ding{55}}%
\newcommand{\Amethods}{A-methods\xspace}
\newcommand{\Bmethods}{B-methods\xspace}
\newcommand{\Dmethods}{D-methods\xspace}
\newcommand{\ellinf}{$\ell_{\infty}$\xspace}

\usepackage{cleveref}
\usepackage{float}

\floatname{algorithm}{Algorithm}

\newcommand{\vct}[1]{\ensuremath{\boldsymbol{#1}}}

\newcommand{\set}[1]{\ensuremath{\mathcal{#1}}}

\newcommand{\mask}{\ensuremath{\vct m}\xspace}
\newcommand{\local}{\ensuremath{L}\xspace}
\newcommand{\gglobal}{\ensuremath{G}\xspace}

\newcommand{\structured}{\ensuremath{S}\xspace}
\newcommand{\unstructured}{\ensuremath{US}\xspace}

\newcommand{\sparsrate}{\ensuremath{s_r}\xspace}

\newcommand{\myparagraph}[1]{\smallskip \noindent \textbf{#1}}
\newcommand{\ie}{{i.e.}\xspace}
\newcommand{\eg}{{e.g.}\xspace}

\begin{document}

\begin{frontmatter}

\title{Adversarial Pruning: A Survey and Benchmark of Pruning Methods for Adversarial Robustness}

\author[unica,sapienza]{Giorgio Piras\texorpdfstring{\corref{mycorrespondingauthor}}{}}
\cortext[mycorrespondingauthor]{Corresponding author}
\ead{giorgio.piras@unica.it}

\author[unica]{Maura Pintor}
\ead{maura.pintor@unica.it}

\author[unica]{Ambra Demontis}
\ead{ambra.demontis@unica.it}

\author[unica]{Battista Biggio}
\ead{battista.biggio@unica.it}

\author[unica,cini]{Giorgio Giacinto}
\ead{giacinto@unica.it}

\author[unige,unica]{Fabio Roli}
\ead{fabio.roli@unige.it}

\affiliation[unica]{organization={University of Cagliari, DIEE}, 
            addressline={Via Marengo 2}, 
            city={Cagliari},
            postcode={09123}, 
            country={Italy}}
\affiliation[sapienza]{organization={Sapienza University of Rome}, 
            addressline={Via Ariosto 25},
            city={Rome},
            postcode={00185},
            country={Italy}}
\affiliation[unige]{organization={University of Genova, DIBRIS}, 
            addressline={Via Dodecaneso 35}, 
            city={Genova},
            postcode={16146},
            country={Italy}}
\affiliation[cini]{organization={CINI, National Cybersecurity Lab, Consorzio Interuniversitario Nazionale per l’Informatica}, 
            addressline={Via Ariosto 25}, 
            city={Roma},
            postcode={00185},
            country={Italy}}


\begin{abstract} 
Recent work has proposed neural network pruning techniques to reduce the size of a network while preserving robustness against adversarial examples, i.e., well-crafted inputs inducing a misclassification.
These methods, which we refer to as \textit{adversarial pruning} methods, involve complex and articulated designs, making it difficult to analyze the differences and establish a fair and accurate comparison.
In this work, we overcome these issues by surveying current adversarial pruning methods and proposing a novel robustness-oriented taxonomy to categorize them based on two main dimensions: the \textit{pipeline}, defining \textit{when} to prune; and the \textit{specifics}, defining \textit{how} to prune. 
We then highlight the limitations of current empirical analyses and propose a novel, fair evaluation benchmark to address them. 
We finally conduct an empirical re-evaluation of current adversarial pruning methods and discuss the results, highlighting the shared traits of top-performing adversarial pruning methods, as well as common issues. We welcome contributions in our publicly-available benchmark at~\url{https://github.com/pralab/AdversarialPruningBenchmark}.
\end{abstract}

\end{frontmatter}

\section{Introduction}
\label{sect:intro}
Deep neural networks tend to be trained in over-parameterized regimes, where the number of model parameters exceeds the dimension of the training set~\cite{belking19-reconciling}. 
Although the performance reached corroborates the need for such a regime, there is still a consistent search for models that suit a resource-constrained scenario, one where the model size cannot be chosen at will but rather minimized.
In addition, it has been shown that although most of these parameters are fundamental to finding a good minimum during training, they then become superfluous~\cite{frankle_lth19}.
It has thus become of great interest to study techniques capable of reducing the network size while still being able to generalize well. 
Compression methods such as pruning~\cite{han15-learning}, quantization~\cite{vanhoucke-nips11} and knowledge distillation~\cite{hinton15-distillation}, have been therefore studied and adapted to current networks. 
In this regard, one popular compression approach is neural network pruning, which originates in the late 80s~\cite{lecun89-obd} and flourishes with the rise of deep learning~\cite{han15-learning,molchanov16-pruning,han16-deepcompression,He_2017_ICCV}. 
The goal of pruning is to reduce the number of network parameters, i.e. the network size, with as little performance degradation as possible. 
 
However, when deployed into security-critical scenarios, neural networks might also be required to be robust against adversarial attacks. In particular, machine-learning models have been discovered to be susceptible to \textit{adversarial examples}, i.e., small perturbations added to the input data that can mislead models' predictions~\cite{biggio13-ecml,szegedy_intriguing_2014}. 
Following the discovery of this phenomenon, an intensive line of research has focused on designing both adversarial attacks~\cite{madry18-iclr, croce20_autoattack} and defenses~\cite{madry18-iclr, zhang19-trades} (\autoref{sect:background}). The robustness of the models against such attacks becomes highly relevant in a wide variety of security-related applications, ranging from self-driving cars to cybersecurity tasks \cite{biggio18_wild}. 

In this regard, when Machine Learning (ML) models need to be both robust against adversarial attacks and compressed, \textit{Adversarial Pruning} (AP) methods stand out by precisely fulfilling this dual need, i.e., creating a pruned network with a given sparsity while preserving robustness. Over the last years, various designs have been used to generate AP methods. In particular, a line of work follows the conventional three-staged pipeline approach proposed in~\cite{han15-learning}, by pruning a robust dense model after pretraining and then finetuning it~\cite{sehwag20-hydra,ye19-radmm}; other approaches amount to prune the model either during~\cite{vemparala21_intrain} or before~\cite{cosentino_search_2019} training; and further work optimizes a robustness-related score assigned to each parameter to decide how to prune \cite{sehwag20-hydra, zhao23-harp}, or solve a robust constrained optimization problem directly~\cite{ye19-radmm,tong22_pwoa}. 
These methods can thus be broadly different in terms of design. 
However, such diversity is hard to contemplate; in fact, the lack of a general framework to categorize state-of-the-art AP methods makes the literature complex and unable to be described in a clear and systematic way. 
Thus, when a new AP method is compared with competing approaches, their design differences tend to be not properly considered, limiting the understanding of each AP design.

In this work, we survey current AP methods and categorize them within a common taxonomy (\autoref{sect:taxonomy}). We find the pruning \textit{pipeline} and the pruning \textit{specifics} to be the two main axes under which the design of an AP method can be described. 
While the pipeline describes when the method prunes the network (e.g., after or during training), the specifics describe how the parameters are removed (e.g., the criterion that defines whether a parameter has to be pruned). 
By detailing the characteristics of each of these axes, we can provide a thorough yet compact description of all AP methods. To the best of our knowledge, we are the first to provide such a categorization for AP methods.

Our work, however, is not limited to providing such a novel categorization. 
As such methods have been tested using different experimental setups and adversarial attacks, yielding different estimates of their accuracy and adversarial robustness, it is difficult to directly compare them and choose the ones that may lead to the best models. To overcome this issue, in this work we design a unified benchmark to re-evaluate the accuracy and adversarial robustness of current AP methods under the same conditions, and using more recent and reliable attack algorithms.  
We thus leverage such benchmark to re-evaluate the existing AP methods and analyze, based on our categorization, the effect of their different designs (\autoref{sect:benchmark}).
With our taxonomy and benchmark, we aim to foster AP methods to be categorically described and fairly tested, thus serving as a ``blueprint" for newly published methods.
We conclude by discussing related work (\autoref{sect:related}),  our contributions, and promising future research directions (\autoref{sect:conclusions}).
\begin{figure}
    \centering
    \includegraphics[width=\linewidth]{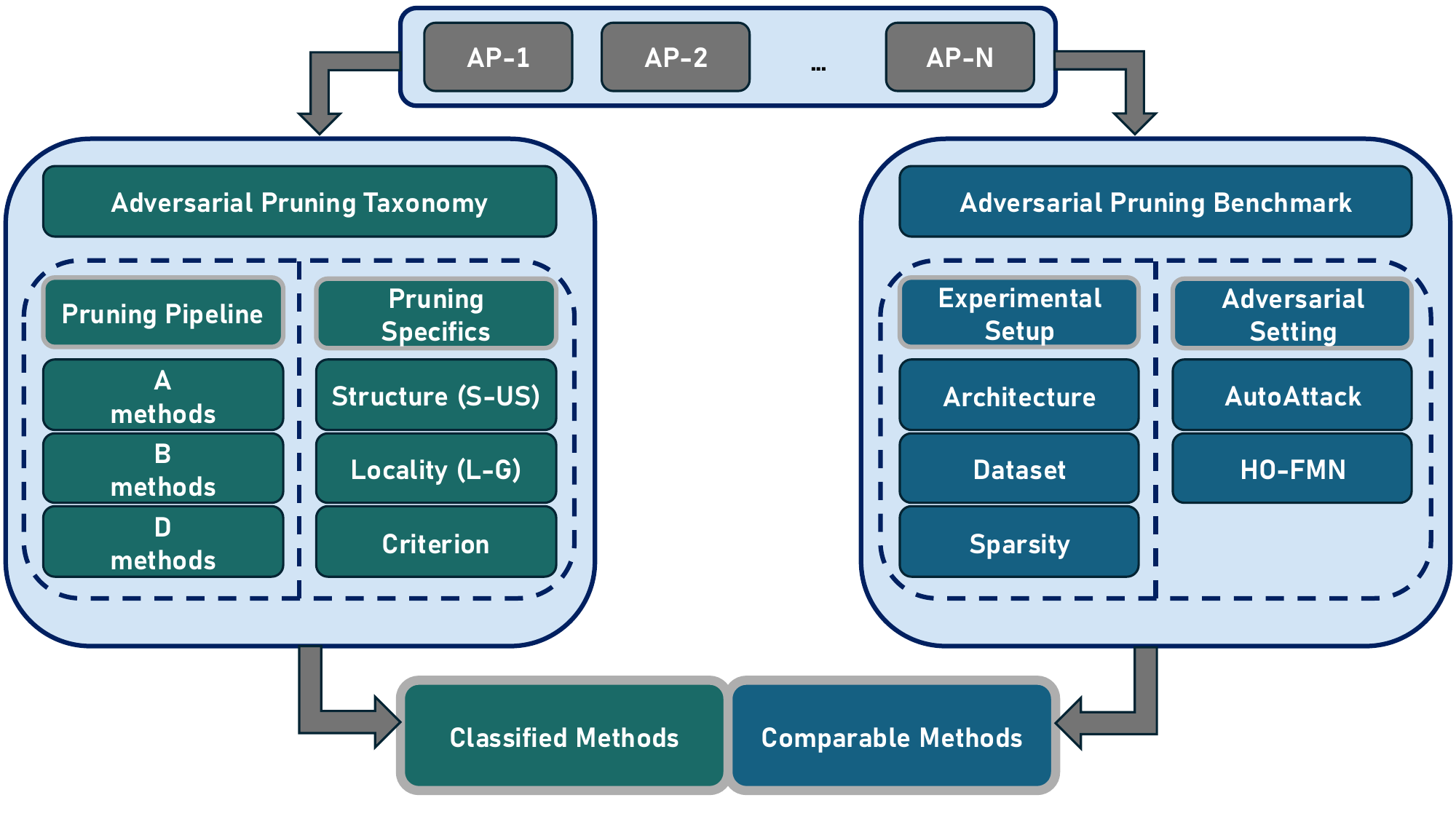}
    \caption{An overview of our taxonomy (left) and our benchmark (right).}
    \label{fig:graphical_abstract}
\end{figure}

\section{Background}\label{sect:background}
This section introduces the basic concepts behind neural network pruning and adversarial robustness. The goal is to provide a clear understanding of pruning and how it can be designed to create sparse networks capable of preserving adversarial robustness.

\subsection{Neural Network Pruning} 
Like many other modern techniques applied to deep learning, neural network pruning originated in the late 80s~\cite{lecun89-obd}. With the rise of deep learning, networks became computationally intensive and memory-requiring, thus often trespassing resource-constrained scenarios where the available memory and computational capability might not be enough to make modern large networks work~\cite{sandler18_mobilenetv2}. Consequently, the research community became increasingly interested in neural network pruning techniques, which found parameters to be fundamental during training and excessive afterward, evolving to highly efficient methods~\cite{han15-learning, molchanov16-pruning, He_2017_ICCV}.

\myparagraph{Pruning formulation.} 
The goal of pruning is to reduce network size, by removing single weights or entire structures (\eg, filters), while not jeopardizing the performance. 
In this formulation, we collectively refer to single weights, entire filters, channels, or kernels as ``parameters," and use the symbol $\vct \theta \in \mathbb R^p$ to denote them, with $p$ being their overall number.
Let us denote with $\set D = (\vct x_i, y_i)_{i=1}^n$ the training set, consisting of $n$ $d$-dimensional samples $\vct x_i \in \set X \subseteq \mathbb R^d$,\footnote{When data is normalized, as for images, typically $\vct x \in \set X = [0, 1]^d$.} along with their class labels $y_i \in \set Y = \{1, \ldots, c\}$. 
Training an ML model amounts to finding a function (within a feasible set, e.g., constrained by a fixed network architecture) that minimizes a loss $\mathcal L(\vct \theta, \vct x, y)$ on the training samples in $\set D$. This typically requires using a convex loss that provides an upper bound on the classification error (i.e., the zero-one loss), while penalizing complex solutions via a regularization term that prevents overfitting. 
The pruning problem starts from a desired sparsity rate $\sparsrate \in [0,1]$, which amounts to retaining only $k= \lfloor p\cdot(1-\sparsrate) \rfloor$ non-zero parameters. We can then formalize it as the problem of finding a binary mask $\vct m \in \{0,1\}^p$ that only retains $k$ parameters while still minimizing the given loss function:
\begin{equation}
\label{eq:pruning_constrained}
\begin{aligned}
\vct m^* \in \underset{\|\vct m\|_0 \leq k}{\arg \min} \, & \sum_{i=1}^n \mathcal{L}(\vct \theta \odot \vct m, \vct x_i, y_i) \, ,
\end{aligned}
\end{equation}
being $\|\vct m\|_0$ the $\ell_0$ norm of the mask, counting its non-zero elements, and $\odot$ the element-wise (Hadamard) product.
We will refer to the algorithms used to solve this problem as $\mathcal C$, being represented by the criterion used to select which parameters to prune or not.

The optimization problem in~\autoref{eq:pruning_constrained} represents a non-convex, combinatorial problem, thus not directly solvable through common approaches such as Stochastic Gradient Descent (SGD). 
For this reason, recent pruning methods have developed different specifics, defining how to prune, as well as different pipelines, defining when to prune. In terms of specifics, multiple approaches have circumvented the problem of~\autoref{eq:pruning_constrained} by using optimizable continuous masks~\cite{sehwag20-hydra}, relaxing the problem constraints~\cite{ye19-radmm}, shrinking the parameters via regularization approaches~\cite{He_2017_ICCV}, or directly pruning the parameters according to naive heuristics, such as pruning the weights with the lowest magnitude (LWM)~\cite{han15-learning}. 
In terms of pipelines, different methods have proposed likewise different instants of the standard training procedure in which to prune. 
In this regard, a classical procedure is represented by the three-staged approach proposed in~\cite{han15-learning}, where the network is pretrained, the pruning mask is applied (\textit{after} training), and the pruned model is then finetuned to restore its performance. 
Different approaches, instead, prune \textit{during} training by jointly creating a mask and updating the network parameters~\cite{huang18-ddsss}, or prune \textit{before} training by finding a sparse subnetwork to be trained from scratch~\cite{frankle_lth19}.

\subsection{Adversarial Robustness}\label{sect:adversarial} 
Following the discovery of learning models being subject to adversarial examples~\cite{biggio13-ecml,szegedy_intriguing_2014}, an important line of research on Adversarial Machine Learning has emerged and evolved over the last years~\cite{biggio18_wild}. 
Research on security in ML has often been addressed as a ``tug of war'' problem: on the one hand, creating powerful attacks to properly assess the vulnerability of ML models; on the other, defending from the very same attacks.

\myparagraph{Attacks.} The ultimate goal of an adversarial attack (\eg, a white box evasion attack)~\cite{biggio13-ecml} is to create an adversarial example that deceives the model and gets misclassified. To accomplish this objective, adversarial attacks can be formulated as optimization problems solving for either the minimum-norm of the perturbation required to achieve misclassification (minimum-norm attacks), or the maximum misclassification confidence possible within a fixed perturbation budget (maximum-confidence attacks). 
In the first case, the problem amounts to finding the smallest perturbation that, when added to the original sample, achieves the misclassification (\eg Fast Minimum-Norm (FMN) attack~\cite{pintor2021fast}). 
In the latter, the problem amounts to finding the maximum confidence possible while the perturbation norm is bounded to a fixed value, such as in the Projected Gradient Descent (PGD) attack~\cite{madry18-iclr}.

\myparagraph{Reliable adversarial evaluation.}
Evaluating the adversarial robustness with a single attack, however, is globally considered to represent a not highly informative and often unreliable evaluation~\cite{carlini19_onevaluating}.
In addition, such attacks are typically run with a default hyperparameter setting, which does not necessarily represent the best attack setting~\cite{mura24_hofmn}. 
Therefore, the recent reference method to evaluate adversarial robustness is represented by AutoAttack, which is a parameter-free ensemble of four different attacks typically improving over single attacks such as PGD~\cite{croce20_autoattack}. 
Similarly, HO-FMN represents an improved version of the FMN attack algorithm by optimizing the attack hyperparameters~\cite{mura24_hofmn}. 
While both attacks enhance the reliability of adversarial evaluations, HO-FMN additionally guarantees to plot \textit{robustness curves} efficiently. 
Robustness curves depict the robustness decrease of a model against an increasing perturbation budget. 
However, while in maximum-confidence attacks (or ensembles, such as AA), a single attack run is bounded by a fixed perturbation budget, in minimum-norm attacks the found adversarial examples have a varying perturbation budget which is found by the attack algorithm, thus requiring a single run to plot the curve (while multiple runs would be required for maximum-confidence attacks).

\myparagraph{Defenses.} Conversely, the goal of a defense is to create a robust model against adversarial attacks. 
The go-to technique, in this case, is represented by Adversarial Training (AT)~\cite{madry18-iclr}, which is formalized as a min-max optimization problem:
\begin{equation}\label{eq:adv_train}
    \min_{\vct \theta} \sum_{i=1}^n \max _{\|\vct \delta_i\|_p \leq \epsilon} \mathcal{L}(\vct \theta, \vct x_i+\vct\delta_i,  y) \, ,
\end{equation}
where the $\ell_p$ norm of the perturbation $\vct \delta$ is upper bounded by $\epsilon$. Note also that $\vct x + \vct \delta \in \set X$, i.e.,  the perturbed sample has to belong to the input space $\set X$. 
The inner maximization finds the worst-case adversarial perturbation, and it is typically solved using adversarial attacks such as PGD~\cite{madry18-iclr}. The outer problem, instead, minimizes the robust loss (computed on the adversarial examples) via SGD, as normally done during training.

\section{Adversarial Pruning}\label{sect:taxonomy}
Following the rise of pruning techniques~\cite{blalock20_state, yeom_21pruning}, the pursuit of robust models led the community to question the effect of pruning on adversarial robustness~\cite{wang_adversarial_2018, guo_sparse_2019, wijayanto_robustness_2018, sehwag19_towards}, thus generating a new spark of research on the interplay between pruning and robustness. 
We thus surveyed more than 50 papers dealing with pruning and robustness and identified a total of 26 proposing novel approaches to create pruned and robust models. 
We labeled these methods as Adversarial Pruning (AP) methods, whose goal is to prune the model to a given \sparsrate while preserving as much robustness as possible.

Initially, AP methods extended standard three-step pipeline with robust objectives~\cite{sehwag19_towards} and used a naive pruning criterion (\eg, LWM). 
However, the research community proposed increasingly complex and diverse designs, employing different robustness-oriented approaches to prune the models while circumventing the problem of~\autoref{eq:pruning_constrained}. 
We, therefore, developed a taxonomy of AP methods, presented in~\autoref{tab:taxonomy}, to systematically classify the APs and have a better understanding of the different designs.
\begin{table*}[htbp]
\caption{The taxonomy of AP methods. We categorize every AP based on pipeline and specifics, and we name the APs with their name/acronym used in the paper. 
Depending on the pipeline, we subdivide the AP methods into 3 major categories: respectively, methods pruning after (\Amethods), before (\Bmethods), and during (\Dmethods) training. 
For each of the three categories, we extend the pipeline with the objective used and the other details, where AT stands for Adversarial Training, NT stands for Natural Training, and n.s. stands for Not Specified (hence, not specifying the objective). We then uniformly identify for each AP method a well-defined set of specifics describing how the pruning is applied, indicating whether Structured (S), Unstructured (US), Local (L), or Global (G) pruning is enabled. Finally, the criterion identifies the rule based on which a mask is created.}
\centering
\resizebox{\textwidth}{!}{%
\setlength{\tabcolsep}{17pt}
\renewcommand{\arraystretch}{1.5}
\begin{tabular}{c|cccl|clclc}
\toprule
\hline
\multicolumn{1}{c|}{} &
  \multicolumn{4}{c|}{\textbf{Pipeline} (\ref{sect:pipeline})} &
  \multicolumn{5}{c}{\textbf{Specifics} (\ref{sect:specifics})} \\ 
\bottomrule
\multicolumn{10}{c}{A-Methods: Pruning \textit{after} Training (\ref{sect:after_methods})} 
\\ 
\toprule
\textbf{Name} &
  \multicolumn{1}{l}{\textbf{Pretraining}} &
  \multicolumn{1}{l}{\textbf{Finetuning}} &
  \multicolumn{2}{c|}{\textbf{1S-IT}} &
  \multicolumn{2}{c}{\textbf{S-US}} &
  \multicolumn{2}{c}{\textbf{L-G}} &
  \multicolumn{1}{l}{\textbf{Criterion}} \\

\multicolumn{1}{l|}{RADMM~\cite{ye19-radmm}} &
  \multicolumn{1}{l}{AT\cite{madry18-iclr}} &
  \multicolumn{1}{l}{AT\cite{madry18-iclr}} &
  \multicolumn{2}{c|}{1S} &
  \multicolumn{2}{c}{S,US} &
  \multicolumn{2}{c}{L} &
  \multicolumn{1}{l}{SOLWM} \\ 
  
\multicolumn{1}{l|}{HYDRA\cite{sehwag20-hydra}} &
  \multicolumn{1}{l}{AT\cite{carmon19-unlabeled}} &
  \multicolumn{1}{l}{AT\cite{carmon19-unlabeled}} &
  \multicolumn{2}{c|}{1S} &
  \multicolumn{2}{c}{US} &
  \multicolumn{2}{c}{L} &
  \multicolumn{1}{l}{LIS} \\
  
\multicolumn{1}{l|}{Heracles\cite{zhao22_heracles}} &
  \multicolumn{1}{l}{AT\cite{madry18-iclr}} &
  \multicolumn{1}{l}{AT\cite{madry18-iclr}}&
  \multicolumn{2}{c|}{1S} &
  \multicolumn{2}{c}{S,US} &
  \multicolumn{2}{c}{G} &
  \multicolumn{1}{l}{LIS} \\
  
\multicolumn{1}{l|}{HARP\cite{zhao23-harp}} &
  \multicolumn{1}{l}{AT\cite{madry18-iclr, zhang19-trades, wang20-mart}} &
  \multicolumn{1}{l}{AT\cite{madry18-iclr, zhang19-trades, wang20-mart}} &
  \multicolumn{2}{c|}{1S} &
  \multicolumn{2}{c}{S,US} &
  \multicolumn{2}{c}{G} &
  \multicolumn{1}{l}{LIS} \\
  
\multicolumn{1}{l|}{PwoA\cite{tong22_pwoa}} &
  \multicolumn{1}{l}{AT\cite{madry18-iclr, zhang19-trades, cui21_lbgat}} &
  \multicolumn{1}{l}{KD\cite{hinton15-distillation}} &
  \multicolumn{2}{c|}{1S} &
  \multicolumn{2}{c}{US} &
  \multicolumn{2}{c}{L} &
  \multicolumn{1}{l}{SOLWM} \\ 
  
\multicolumn{1}{l|}{MAD\cite{lee_masking_2022}} &
  \multicolumn{1}{l}{AT\cite{madry18-iclr}} &
  \multicolumn{1}{l}{AT\cite{madry18-iclr,wong20_fastat}} &
  \multicolumn{2}{c|}{1S} &
  \multicolumn{2}{c}{US} &
  \multicolumn{2}{c}{G} &
  \multicolumn{1}{l}{LIS} \\
  
\multicolumn{1}{l|}{Sehwag19\cite{sehwag19_towards}} &
  \multicolumn{1}{l}{AT\cite{madry18-iclr}} &
  \multicolumn{1}{l}{AT\cite{madry18-iclr}} &
  \multicolumn{2}{c|}{IT} &
  \multicolumn{2}{c}{S,US} &
  \multicolumn{2}{c}{L} &
  \multicolumn{1}{l}{LWM} \\
  
\multicolumn{1}{l|}{RSR\cite{rakin19_robust}} &
  \multicolumn{1}{l}{AT\cite{madry18-iclr} + CNI\cite{he19_cni}}&
  \multicolumn{1}{l}{n.s.} &
  \multicolumn{2}{c|}{1S} &
  \multicolumn{2}{c}{US} &
  \multicolumn{2}{c}{G} &
  \multicolumn{1}{l}{RELWM} \\ 
  
\multicolumn{1}{l|}{BNAP\cite{wei_batch_2021}} &
  \multicolumn{1}{l}{AT\cite{madry18-iclr}} &
  \multicolumn{1}{l}{AT\cite{madry18-iclr}} &
  \multicolumn{2}{c|}{1S,IT} &
  \multicolumn{2}{c}{S,US} &
  \multicolumn{2}{c}{G} &
  \multicolumn{1}{l}{LIS} \\
  
\multicolumn{1}{l|}{RFP\cite{lim_robustness-aware_2021}} &
  \multicolumn{1}{l}{AT\cite{ilyas19_bugs}} &
  \multicolumn{1}{l}{AT\cite{ilyas19_bugs}} &
  \multicolumn{2}{c|}{1S} &
  \multicolumn{2}{c}{S} &
  \multicolumn{2}{c}{L} &
  \multicolumn{1}{l}{HGM} \\
  
\multicolumn{1}{l|}{Deadwooding\cite{kaur_deadwooding_2022}} &
  \multicolumn{1}{l}{n.s.} &
  \multicolumn{1}{l}{KD\cite{hinton15-distillation}+AT\cite{goodfellow15-iclr}} &
  \multicolumn{2}{c|}{1S} &
  \multicolumn{2}{c}{US} &
  \multicolumn{2}{c}{G} &
  \multicolumn{1}{l}{SOLWM} \\ 
  
\multicolumn{1}{l|}{FRE\cite{zhuang23_adversarial}} &
  \multicolumn{1}{l}{AT\cite{zhang19-trades}} &
  \multicolumn{1}{l}{AT\cite{zhang19-trades}} &
  \multicolumn{2}{c|}{IT} &
  \multicolumn{2}{c}{S} &
  \multicolumn{2}{c}{G} &
  \multicolumn{1}{l}{LIS} \\
  
\multicolumn{1}{l|}{Luo23\cite{luo23_towards}} &
  \multicolumn{1}{l}{AT\cite{madry18-iclr}} &
  \multicolumn{1}{l}{AT\cite{madry18-iclr}} &
  \multicolumn{2}{c|}{1S} &
  \multicolumn{2}{c}{S} &
  \multicolumn{2}{c}{L} &
  \multicolumn{1}{l}{LIS} \\
  
\multicolumn{1}{l|}{SR-GKP\cite{zhong2023adv_robust_gkp}} &
  \multicolumn{1}{l}{NT} &
  \multicolumn{1}{l}{NT} &
  \multicolumn{2}{c|}{1S} &
  \multicolumn{2}{c}{S} &
  \multicolumn{2}{c}{L} &
  \multicolumn{1}{l}{LIS} \\
  
\multicolumn{1}{l|}{FSRP\cite{qian_robust_2023}} &
  \multicolumn{1}{l}{AT\cite{madry18-iclr, zhang19-trades, wang20-mart}} &
  \multicolumn{1}{l}{AT\cite{madry18-iclr, zhang19-trades, wang20-mart}} &
  \multicolumn{2}{c|}{1S} &
  \multicolumn{2}{c}{S} &
  \multicolumn{2}{c}{L} &
  \multicolumn{1}{l}{LIS} \\ 

\bottomrule
\multicolumn{10}{c}{B-Methods: Pruning \textit{before} Training (\ref{sect:before_methods})} \\ 
\toprule

\textbf{Name} &
  \multicolumn{1}{l}{\textbf{Pruning step}} &
  \multicolumn{1}{l}{\textbf{Training step}} &
  \multicolumn{2}{c|}{\textbf{1S-IT}} &
  \multicolumn{2}{c}{\textbf{S-US}} &
  \multicolumn{2}{c}{\textbf{L-G}} &
  \multicolumn{1}{l}{\textbf{Criterion}} \\

\multicolumn{1}{l|}{Cosentino19\cite{cosentino_search_2019}} &
\multicolumn{1}{l}{AT\cite{madry18-iclr, goodfellow15-iclr}}& 
\multicolumn{1}{l}{AT\cite{madry18-iclr, goodfellow15-iclr}}& 
\multicolumn{2}{c|}{IT} &
  \multicolumn{2}{c}{US} & 
  \multicolumn{2}{c}{G} &
  \multicolumn{1}{l}{LWM} \\ 

\multicolumn{1}{l|}{Li20\cite{li_towards_2020}} &
\multicolumn{1}{l}{AT\cite{goodfellow15-iclr}}& 
\multicolumn{1}{l}{AT\cite{madry18-iclr}}&
\multicolumn{2}{c|}{1S} &
  \multicolumn{2}{c}{US} &
  \multicolumn{2}{c}{G} &
  \multicolumn{1}{l}{LWM} \\ 
  
\multicolumn{1}{l|}{Wang20\cite{wang_achieving_2020}} &
\multicolumn{1}{l}{AT\cite{madry18-iclr}}& 
\multicolumn{1}{l}{AT\cite{madry18-iclr}}& 
\multicolumn{2}{c|}{IT} &
  \multicolumn{2}{c}{US} &
  \multicolumn{2}{c}{G} &
  \multicolumn{1}{l}{LWM} \\ 

\multicolumn{1}{l|}{RST\cite{fu21_rst}} &
\multicolumn{1}{l}{AT\cite{madry18-iclr}}& 
\multicolumn{1}{l}{None}&
\multicolumn{2}{c|}{1S} &
  \multicolumn{2}{c}{US} &
  \multicolumn{2}{c}{L} &
  \multicolumn{1}{l}{LIS} \\ 

\multicolumn{1}{l|}{RobustBird\cite{chen_sparsity_2021}} &
\multicolumn{1}{l}{AT\cite{madry18-iclr}}& 
\multicolumn{1}{l}{AT\cite{madry18-iclr}}& 
\multicolumn{2}{c|}{IT} &
  \multicolumn{2}{c}{US} &
  \multicolumn{2}{c}{G} &
  \multicolumn{1}{l}{LWM} \\
  
\multicolumn{1}{l|}{AWT\cite{shi22_finding}} &
\multicolumn{1}{l}{AT\cite{madry18-iclr}}& 
\multicolumn{1}{l}{AT\cite{madry18-iclr}}& 
\multicolumn{2}{c|}{n.s.} &
  \multicolumn{2}{c}{US} &
  \multicolumn{2}{c}{n.s.} &
  \multicolumn{1}{l}{LWM} \\\hline

\bottomrule
\multicolumn{10}{c}{D-Methods: Pruning \textit{during} Training (\ref{sect:during_methods})} 
\\ 
\toprule

\textbf{Name} &
  \multicolumn{4}{l|}{\textbf{Training step}} &
  \multicolumn{2}{c}{\textbf{S-US}} &
  \multicolumn{2}{c}{\textbf{L-G}} &
  \multicolumn{1}{l}{\textbf{Criterion}} \\

\multicolumn{1}{l|}{TwinRep\cite{li_twinrep_2023}} & 
\multicolumn{4}{l|}{AT\cite{madry18-iclr}}&
  \multicolumn{2}{c}{S,US} &
  \multicolumn{2}{c}{G} &
  \multicolumn{1}{l}{RELWM} \\ 
  
\multicolumn{1}{l|}{BCS-P\cite{ozdenizci_training_2021}} & 
\multicolumn{4}{l|}{AT\cite{madry18-iclr,zhang19-trades, wang20-mart, carmon19-unlabeled,kurakin17_iclr}} &
  \multicolumn{2}{c}{US} &
  \multicolumn{2}{c}{L,G} &
  \multicolumn{1}{l}{BCS} \\
  
\multicolumn{1}{l|}{DNR\cite{kundu_tunable_2020}} &
\multicolumn{4}{l|}{AT\cite{madry18-iclr}}&
  \multicolumn{2}{c}{S,US} &
  \multicolumn{2}{c}{G} &
  \multicolumn{1}{l}{SOLWM} \\ 
  
\multicolumn{1}{l|}{InTrain\cite{vemparala21_intrain}} &
\multicolumn{4}{l|}{AT\cite{wong20_fastat}}&
  \multicolumn{2}{c}{S,US} &
  \multicolumn{2}{c}{G} &
  \multicolumn{1}{l}{LIS} \\ 
  
\multicolumn{1}{l|}{FlyingBird\cite{chen_sparsity_2021}} &
\multicolumn{4}{l|}{AT\cite{madry18-iclr}}&
  \multicolumn{2}{c}{US} &
  \multicolumn{2}{c}{G} &
  \multicolumn{1}{l}{LWM} \\\hline 

  \bottomrule
\end{tabular}%
}
\label{tab:taxonomy}
\end{table*}

\subsection{Adversarial Pruning Taxonomy} 
We defined two main pillars through which AP methods can be classified, pruning \textbf{pipeline} and pruning \textbf{specifics}. The two pillars answer two questions of paramount importance: respectively, \textit{when to prune} and \textit{how to prune}. 
The pipeline, described in~\autoref{sect:pipeline}, precisely indicates the time at which the pruning mask is applied with respect to the training stage. 
Depending on the pipeline, we identified three major categories of AP methods: pruning after training (\Amethods), pruning before training (\Bmethods), and pruning during training (\Dmethods). 
Based on this major categorization, each method defines a set of robust objectives and details employed in the pipeline.
The specifics instead, described in~\autoref{sect:specifics}, define how the pruning is applied and serve as a scheme for the mask design. 
We identify, as specifics, the pruning structure, locality, and the pruning criteria.

\subsection{Pruning Pipeline}\label{sect:pipeline} 
Our pruning pipeline formulation can be defined as a \textit{two-level} categorization: the first level defines when the mask is created and applied with respect to the training step: we thus define three major categories described as \Amethods (~\autoref{sect:after_methods}), \Bmethods (~\autoref{sect:before_methods}), and \Dmethods (~\autoref{sect:during_methods}). The second level, instead, defines in more detail the robust objectives used in the pipeline (e.g., adversarial training) and the approaches used in each pipeline (which can vary depending on the first level).

\subsubsection{A-Methods: Pruning after Training}\label{sect:after_methods} 
\Amethods, follow the conventional three-step pipeline described in~\cite{han15-learning} extended with robust objectives such as Adversarial Training (AT). 
Hence, models are pretrained, pruned, and finetuned with robust objectives.
In this pipeline configuration, the mask \mask is therefore applied to the dense model after the training step. 

\myparagraph{Pretraining and finetuning.} 
The extension of the pipeline in~\cite{han15-learning} to adversarial robustness simply requires using robust training objectives. 
Therefore, within \Amethods, we define both pretraining and finetuning objectives.
Unsurprisingly, given its wide use, we find PGD AT~\cite{madry18-iclr} to be the most commonly employed pretraining and finetuning objective. Every AP method following such pipeline uses AT for both pretraining and finetuning, except for the cases of SR-GKP~\cite{zhong2023adv_robust_gkp}, where natural training (NT) is used for both steps, and PwoA~\cite{tong22_pwoa}, where the pruned model in the finetuning step distills knowledge from the teacher pretrained robust model.

\myparagraph{1S-IT.} We refer to one-shot (1S) pruning when the pruning and finetuning steps are run once each, thus reaching the desired sparsity \sparsrate in one shot. On the other hand, we refer to iterative (IT) pruning when the pruning and finetuning steps are run more than once, thus reaching the desired \sparsrate progressively (e.g., 3 iterations pruning 30\% of parameters to reach a desired 90\% \sparsrate). Besides the seminal work in~\cite{sehwag19_towards}, which was based on IT, most of the more recent work uses a 1S approach, typically avoiding increasing the computational burden. The analysis in multiple work comparing the two versions of \Amethods pipelines shows negligible differences between 1S and IT~\cite{kaur_deadwooding_2022, sehwag20-hydra}. 

\subsubsection{B-Methods: Pruning before Training}\label{sect:before_methods}
\Bmethods search a sparse subnetwork which is then trained in isolation to match the dense performances, following the widely known Lottery Ticket Hypothesis (LTH~\cite{frankle_lth19}). 
In this case, the pipeline is summarized in two steps: pruning and training. The pruning step involves a search for the \textit{winning ticket}, which consists of preliminary training and pruning of the network, resulting in a subnetwork that, trained in isolation, matches the dense performances. The training step, instead, consists in isolating the pruned network with the original starting random initialization (or, in some cases, initializing with a new random initialization~\cite{cosentino_search_2019}) and then training the resulting subnetwork from scratch. 
We specify that the watershed in labeling methods conforming to such pipeline as \Bmethods opposed to \Amethods is their rationale, which posits that there exist subnetworks that, when trained in isolation, can match the performances of their dense counterpart.  
Therefore, we consider the main training step of the network to be applied on the subnetwork and, consequently, the pruning being applied before training. 
We thus consider the first step of the pipeline as a simple architecture search.
Extending from~\cite{frankle_lth19}, LTH has been validated on robust models by incorporating, starting from~\cite{cosentino_search_2019}, robust objectives in both pruning and training steps.

\myparagraph{Pruning step.} The pruning step in \Bmethods is represented by a sparse architecture search. The procedure starts from a dense, randomly initialized model. The model is then trained for a few iterations to be iteratively pruned (IT) or pruned after the iterations in one-shot (1S). \Bmethods use a robust objective such as AT~\cite{madry18-iclr} within the search, leading to robust winning tickets~\cite{cosentino_search_2019, li_towards_2020, wang_achieving_2020, fu21_rst, shi22_finding}, even in the early training stages~\cite{chen_sparsity_2021}.  

\myparagraph{Training step.} The training step consists in training from scratch the subnetwork found in the first pruning step, yet starting from the original initialization of the dense model (or a new random one). While the majority of the APs classified as \Bmethods use AT as a training procedure, the case of RST~\cite{fu21_rst} stands out: in RST, the pruning stage consists in a mask search on the randomly initialized model that leads to a robust ticket. This ticket is considered to be robust from scratch and does not need further training. 

\myparagraph{1S-1T.} The one-shot (1S) or iterative (IT) approaches define when the final mask is obtained. While classic LTH~\cite{frankle_lth19} adopts an IT approach, several AP methods opt for 1S pruning with a single run~\cite{li_towards_2020, fu21_rst}. The analysis in~\cite{li_towards_2020}, once again, shows negligible differences between 1S and IT approaches from \Bmethods. 

\subsubsection{D-Methods: Pruning during Training}\label{sect:during_methods}
\Dmethods represent end-to-end pipelines by jointly learning a mask used to prune the model and the networks' parameters. Therefore, such methods need a single training run and a robust objective, representing an efficient compromise. 
Following also seminal work~\cite{dettmers19_scratch}, \Dmethods use a dynamic approach by pruning and regrowing parameters (i.e., restoring previously pruned parameters to prune new ones while keeping the same sparsity rate) during robust training. In this way, the sparsity constraint \sparsrate can be guaranteed throughout the training run by compensating pruning with regrowing (i.e., if a new parameter has to be pruned, a new one has to be regrown). Such procedure implies that \Dmethods decouple their pipeline from a classical 1S/IT notation. 

\myparagraph{Training step.} The \Dmethods training step is typically more demanding, since the pruning procedure is jointly run throughout the training and includes regrowing the weights. 
In any case, the \Dmethods are all found to be subject to AT~\cite{madry18-iclr} approaches just like the other kinds of pipelines. 

\subsection{Specifics}\label{sect:specifics}
Creating a binary mask \mask, depends upon a set of specifics that are associated with pruning: (i) the structure, which defines the kind of parameters that are removed (e.g., single weights or entire filters)~\cite{han15-learning, haocvpr16-filters}; (ii) the locality, which defines whether the desired sparsity rate \sparsrate is accomplished locally in each layer or globally~\cite{blalock20_state}; and (iii) the pruning criteria, which is responsible for identifying the parameters to be removed, and has been designed in different ways to circumvent the problem of~\autoref{eq:pruning_constrained}. 
While structure and locality are independent of the goal aspiring to reach in the sparse network (e.g., robustness for APs), the pruning criterion can be designed in different ways and has thus been geared towards preserving adversarial robustness in AP methods.

\subsubsection{Structure (S vs US)}\label{sect:structure}  
When removing single weights, pruning is said to be unstructured (\unstructured), while when removing entire structures (\eg filters), pruning is referred to as structured (\structured)~\cite{haocvpr16-filters, molchanov16-pruning, He_2017_ICCV, blalock20_state}. 
In the literature, \structured pruning is often considered more relevant than \unstructured, since removing entire structures allows creating a lighter architecture; \unstructured pruning instead requires dedicated hardware to be correctly exploited~\cite{liu2023lessons}. 
In fact, removing entire structures from a dense network equals directly reducing size, which can thus allow to be directly exploited. 
On the other hand, removing single weights enables a network with a sparse pattern, which can only yield a limited speedup on regular hardware.
However, \unstructured pruning enables a higher degree of flexibility and allows the preservation of higher performances than structured pruning when compared to the same level of sparsity. 
In addition, \unstructured it is easier to implement and represents a highly useful mathematical prototype and test bench while getting increasing support for practical applications~\cite{liu2023lessons}.
Among the considered AP methods, 21/26 implement \unstructured and 13/26 implement \structured.

\subsubsection{Locality (L vs G)}\label{sect:locality} 
When pruning is applied locally (\local), the desired \sparsrate is reached equally in each layer, while when applied globally (\gglobal), the sparsity \sparsrate is reached globally, thus treating the parameters in each layer equally and pruning them independently from their belonging layer.
Therefore, when pruning is \local, the rule used to prune the parameters (i.e., the pruning criterion of~\autoref{sect:criterion}) is applied with the same \sparsrate in each layer, while when pruning is \gglobal it is extended to the whole network. 
Consequently, while in \local pruning each layer has the same sparsity equal to \sparsrate, in \gglobal pruning each layer can have different sparsities that, summed up, satisfy the desired \sparsrate globally. 
Interestingly, methods such as HARP~\cite{zhao23-harp} and FlyingBird~\cite{chen_sparsity_2021}, that allow for different sparsities among layers and are thus labeled as \gglobal, control and optimize the sparsity rate of each layer.
Similarly, DNR~\cite{kundu_tunable_2020} extends from RADMM and allows layers to adjust their sparsity dynamically.
Among the AP methods, a total of 11/26 methods implement local pruning, while 15/26 adopt a global approach. 

\subsubsection{Pruning Criteria}\label{sect:criterion}
The pruning criterion defines a rule based on which a parameter is pruned or not. 
In essence, this translates to defining the algorithm $\mathcal{C}$.
The most classical criterion is represented by the Lowest Weight Magnitude (LWM) based pruning~\cite{han15-learning}, which removes the weights with the lowest magnitude based on the assumption that they tend to have the least harmful impact on the loss. 
An AP method, however, aims to prune the parameters that cause the least drop in robustness. 
Consequently, many of the work in~\autoref{tab:taxonomy} extended the problem of~\autoref{eq:pruning_constrained} to the adversarial case. 
However, independently from the objective, the sparsity constraint still makes the problem non-convex and combinatorial: this led AP methods to adopt multiple solutions circumventing this limitation in different ways while attempting to preserve robustness.

\myparagraph{SOlver-based Lowest Weight Magnitude (SOLWM).} The SOLWM criterion represents AP methods whose pruning criterion's primary goal is to first solve the robust constrained optimization problem imposed by sparsity (hence creating a robust model while satisfying the desired \sparsrate) and then prune the parameters with the lowest magnitude. 
Many of the collected AP methods resort to the \textit{alternating direction method of multipliers} (ADMM) as a solver for the constrained optimization problem~\cite{ye19-radmm, kundu_tunable_2020, tong22_pwoa}. 
ADMM is an optimization algorithm that, via the definition of an augmented Lagrangian, decomposes the main non-convex combinatorial problem into two sub-problems solved iteratively by SGD. 
In both RADMM~\cite{ye19-radmm} and PwoA~\cite{tong22_pwoa}, ADMM is applied on a robust pretrained network before finetuning, while in the modified formulation of DNR~\cite{kundu_tunable_2020}, the optimization is leveraged during training. 
Alternatively to ADMM, we identified Deadwooding~\cite{kaur_deadwooding_2022} relaxing the constraints using Lagrangian multipliers and proposing a \textit{Lagrangian dual method} to solve the optimization problem.
When pruning using a SOLWM criterion, the training process designed to circumvent the constraint equals regularizing the parameters with a higher penalty when their contribution to the robust loss is higher (thus shrinking their values). 
Therefore, as a final step, such methods prune the parameters with the lowest magnitude (hence, with LWM).

\myparagraph{REgularization-based Lowest Weight Magnitude (RELWM).} This criterion represents AP methods explicitly regularizing the network parameters, which reduces their magnitude, and then pruning the weights with the lowest magnitude itself. By having a generally low weight magnitude and pruning with LWM, the harm induced by pruning on adversarial robustness is believed to be reduced. We identified a total of two AP methods adopting such an approach: RSR~\cite{rakin19_robust} uses an L\textsubscript{1} penalty term (i.e., lasso) to encourage sparsity in the network in the adversarial pretraining step, and then prunes the weights with the least magnitude; TwinRep~\cite{li_twinrep_2023}, reparameterizes the original network with the element-wise product of two matrices (i.e., $\vct \theta = \vct W_1 \odot \vct W_2$) and learns both matrices during training. The overall effect of this procedure is similar to adding a weight decay factor, i.e., reducing the magnitude of the weights and facilitating pruning them with a lower impact on the overall performance.

\myparagraph{Least Importance Score (LIS).} Pruning, according to the LIS, entails defining, for each parameter, a score based on which to prune. 
We define a score as an induced measure defining how important the parameter is for the overall network robustness, thus not considering directly accessible measures such as the weight magnitude for which naive pruning criteria are already defined. 
As shown in~\autoref{tab:taxonomy}, LIS is the most common criterion for AP methods. 
However, the way in which importance is defined varies profoundly among AP methods. 
The approaches in HYDRA~\cite{sehwag20-hydra}, HARP~\cite{zhao23-harp}, InTrain~\cite{vemparala21_intrain}, RST~\cite{fu21_rst} and Heracles~\cite{zhao22_heracles}, all define a continuous mask (i.e., a set of importance scores of equal dimension to the parameters matrix) in an empirical risk minimization problem. 
The mask is then optimized using a robust objective, and the weights corresponding to the lower importance score (i.e., the ones with the least impact on adversarial robustness) are finally pruned. 
Other APs, formulate the importance differently: MAD~\cite{lee_masking_2022} computes an estimate of the adversarial saliency for each parameter and prunes the least salient ones, as done for filters, similarly, in~\cite{zhuang23_adversarial}; FSRP~\cite{qian_robust_2023}, prunes the filters based on the high-frequency components of their feature maps based on the rationale that adversarial examples mainly exist within such frequencies; similarly, in BNAP~\cite{wei_batch_2021}, the scaling factor of batch normalization layers is found to be representative of these components, being thus multiplied to the weights to define importance. 
Every AP method using LIS creates a binary mask based on the importance obtained and the desired \sparsrate, which is then multiplied to the parameters to prune.

\myparagraph{Other pruning criteria.}
We identified multiple AP methods resorting to naive criteria such as LWM~\cite{sehwag19_towards, chen_sparsity_2021}. Using the LWM approach is also particularly common for \Bmethods (pruning before training), where the pipeline for the ticket search is typically prioritized more than the criterion, such as in~\cite{li_towards_2020, wang_achieving_2020, chen_sparsity_2021, shi22_finding}. We additionally find RFP~\cite{lim_robustness-aware_2021} employing the Highest Gradient Magnitude (HGM) for filter pruning. Finally, differently from all other APs, BCS-P~\cite{ozdenizci_training_2021} optimizes the network with a negative log-posterior loss combining a sparsity prior with a robust training objective. Then, the network parameters are sampled from the posterior distribution.
For regrowing in \Dmethods instead, the usual approach is to regrow based on the inverse of the pruning criteria, such as for InTrain~\cite{vemparala21_intrain}. Similarly, TwinRep~\cite{li_twinrep_2023} reparameterizes directly with the product of the two matrices, while BCS-P~\cite{ozdenizci_training_2021} simply resamples the parameters. FlyingBird~\cite{chen_sparsity_2021} instead, uses the Highest Gradient Magnitude (HGM), while DNR~\cite{kundu_tunable_2020}, following~\cite{dettmers19_scratch}, uses normalized momentum.

\subsection{Standard vs Adversarial Pruning Methods}
After presenting the taxonomy of AP methods, it is fundamental to clarify the difference between standard and AP methods. Following the general pruning formulation in~\autoref{eq:pruning_constrained}, we identify an objective (i.e., minimizing the loss), and a constraint imposed by pruning to obtain sparsity. While the constraint is the same, the objective clearly changes from standard to AP methods, setting a different goal. Hence, while in standard pruning we will simply minimize the loss $\mathcal{L}$ computed on clean examples, APs consider a robust loss, following~\autoref{eq:adv_train}: 
\begin{equation}
\label{eq:adversarial_pruning_constrained}
\begin{aligned}
\vct m^* \in \underset{\|\vct m\|_0 \leq k}{\arg \min} \sum_{i=1}^n \max _{\|\vct \delta_i\|_p \leq \epsilon} \mathcal{L}(\vct \theta \odot \vct m, \vct x_i+\vct\delta_i,  y) \, ,
\end{aligned}
\end{equation}
which indicates that the mask $\vct m^*$ will result from minimizing the loss computed on the adversarial example $\vct x + \vct \delta$, following~\autoref{eq:adv_train}. 
This difference can be encountered, through our taxonomy, by observing the objectives employed in the pipelines, which mostly resort to AT techniques. Similarly, the AP criteria presented in~\autoref{sect:criterion} explicitly incorporate robustness into their design: e.g., LIS prunes based on the importance of the parameter with respect to adversarial robustness; SOLWM solves the robust constrained optimization problem; RELWM encourages sparsity while enhancing robustness. 
However, while the objective is discriminant, both standard and AP methods prune weights. Hence, while some details of the pipelines and specifics are unique to the kind of pruning, some other simply abstract from it; e.g., both standard and AP define locality and structure, and considering the sub-optimality of na\"ive pruning criteria for both tasks, both standard and AP methods can be found to use LWM.

\begin{figure*}[ht]
    \centering
    \includegraphics[width=0.9\linewidth]{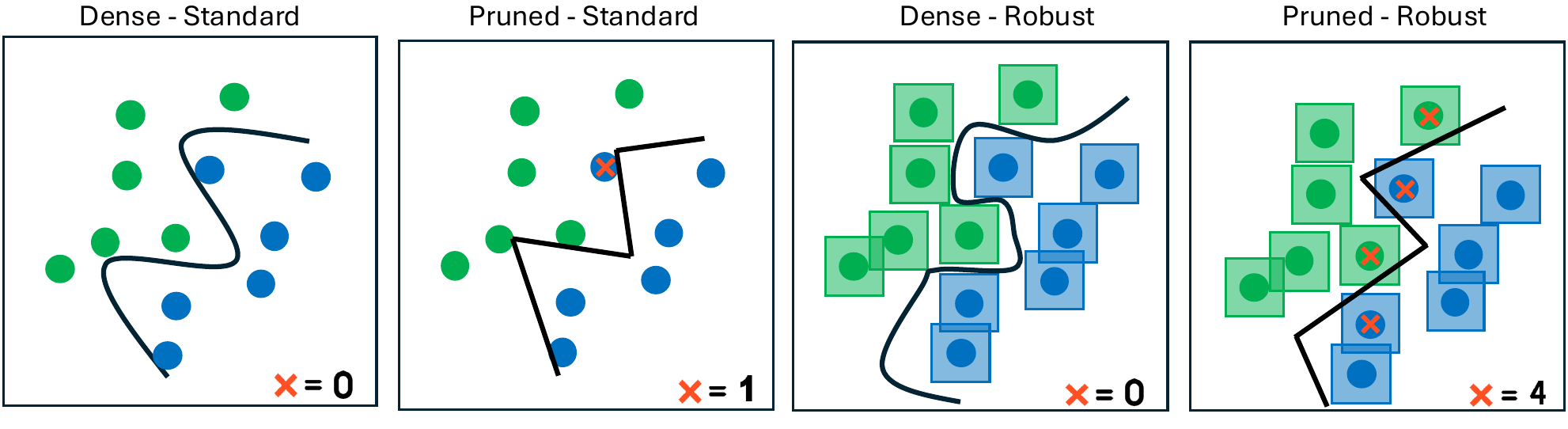}
    \caption{A representation of the boundary complexity for different model-task pairs, where red crosses indicate a misclassified or possibly adversarial example. When the model is pruned and the task is standard accuracy, creating a complex decision boundary becomes harder compared to a dense model (first and second figures). When the task is robustness, such difficulty escalates as the boundary is ``trained" including adversarial examples within a given perturbation norm (e.g., an \ellinf constraint).}
    \label{fig:boundary_pruning}
\end{figure*}

\myparagraph{The challenge of AP methods.}
Finally, let us remark a major challenge of AP methods compared to standard pruning. 
Considering adversarial training  from~\autoref{eq:adv_train}, and as also highlighted in~\cite{madry18-iclr}, building a robust model requires designing a much more complex decision boundary. As we prune a model, and reduce the number of parameters, retaining a complex boundary becomes harder. Thus, while performing AT is challenging per se, on a pruned model the challenge escalates, as represented in~\autoref{fig:boundary_pruning}. 

\section{Benchmarking Adversarial Pruning}\label{sect:benchmark}
In~\autoref{sect:taxonomy}, we classified the AP methods based on pipeline and specifics, and then analyzed the core disparities. 
Upon a categorization of the methods, we took on the challenge of comparing each AP's robustness while acknowledging the proposed taxonomy. 
However, our investigation revealed the impracticability of directly deriving such a comparison due to the many differences in the experimental setups of these evaluations. 
In fact, newly published APs are compared on equal setups only to a few of the ``top-notch" AP methods, and while such analysis can be sufficient to assess the AP novelty, it is not enough to compare the robustness of \emph{each} AP method (i.e., our goal).
Furthermore, we find multiple issues associated with the adversarial evaluations conducted in these works, which represent a well-known problem in the field of adversarial ML, ultimately leading to overestimating the robustness of the models~\cite{carlini19_onevaluating, pintor22_indicators}.
All these obstacles are implicit limitations in the absence of a benchmark.
In this section, we thus present our benchmark implementation and re-evaluate the pruned models produced by AP methods with a \textit{comparable} experimental setup and \textit{fair} adversarial setting.
In this regard, we first bring to light, in~\autoref{sect:evaluations}, a number of problems related to the current AP evaluations, which are often found to be below par with respect to previous work. 
We then present the benchmark experimental settings in~\autoref{sect:exp_settings} and re-evaluate the available AP methods in~\autoref{sect:results}, where we conclude by discussing the re-evaluation results with particular attention to our taxonomy.

\subsection{Adversarial Pruning Evaluations}\label{sect:evaluations}
\begin{table}[t]
\caption{AP evaluations. AA and PGD refer to AP evaluations using AutoAttack~\cite{croce20_autoattack} and PGD~\cite{madry18-iclr}, respectively. Additional attacks are reported under ``Others", including the Fast Gradient Sign Method attack (FGSM), Carlini\&Wagner (CW), Zeroth Order Optimization (ZOO), DeepFool and Brendel \& Bethge (BB)~\cite{cina24attackbench}. In ``Robustness curve" we report whether multiple perturbation sizes were used for the given attacks, while through ``Iter$>$10" we report whether the PGD attacks were run with more than 10  iterations.}
\centering
\resizebox{0.5\textwidth}{!}{%
\begin{tabular}{@{}ccccccc@{}}
\toprule
\textbf{Name} & \textbf{AA} & \textbf{PGD} & \textbf{Others} & \textbf{Robustness Curve} & \textbf{Iter$>$10}\\ \midrule
\multicolumn{1}{l|}{RADMM\cite{ye19-radmm}}         & \xmark                     & \checkmark & CW           & \xmark & \checkmark \\
\multicolumn{1}{l|}{HYDRA\cite{sehwag19_towards}}          & \checkmark                     & \checkmark & \xmark           & \checkmark  & \checkmark \\
\multicolumn{1}{l|}{{Heracles\cite{zhao22_heracles}}} & \xmark                     & \checkmark & FGSM, CW     & \xmark  & \checkmark  \\
\multicolumn{1}{l|}{HARP\cite{zhao23-harp}}           & \checkmark                    & \checkmark & CW           & \xmark & \checkmark  \\
\multicolumn{1}{l|}{PwoA\cite{tong22_pwoa}}           & \checkmark                    & \checkmark & FGSM, CW     & \xmark &  \checkmark  & \\
\multicolumn{1}{l|}{MAD\cite{lee_masking_2022}}            & \checkmark                    & \checkmark & FGSM, CW     & \xmark &  \checkmark  \\
\multicolumn{1}{l|}{{Sehwag19\cite{sehwag19_towards}}} & \xmark                     & \checkmark & \xmark           & \xmark & n.s.  \\
\multicolumn{1}{l|}{{RSR\cite{rakin19_robust}}}      & \xmark                     & \checkmark & FGSM ZOO     & \xmark &  \xmark \\
\multicolumn{1}{l|}{{BNAP\cite{wei_batch_2021}}}     & \xmark                     & \checkmark & \xmark           & \xmark & \checkmark  \\
\multicolumn{1}{l|}{{RFP\cite{lim_robustness-aware_2021}}}      & \xmark                     & \checkmark & \xmark           & \xmark  & \checkmark \\
\multicolumn{1}{l|}{Deadwooding\cite{kaur_deadwooding_2022}}    & \checkmark                    & \checkmark & DeepFool & \xmark & \checkmark \\
\multicolumn{1}{l|}{{FRE\cite{zhuang23_adversarial}}}      & \xmark                     & \checkmark & \xmark           & \xmark  & \xmark \\
\multicolumn{1}{l|}{Luo23\cite{luo23_towards}}          & \xmark & \checkmark & \xmark           & \xmark & \checkmark  \\
\multicolumn{1}{l|}{{SR-GKP\cite{zhong2023adv_robust_gkp}}}   & \xmark                     & \checkmark & FGSM         & \xmark & \xmark \\
\multicolumn{1}{l|}{FSRP\cite{qian_robust_2023}}           & \checkmark                    & \checkmark & FGSM         & \xmark   & \checkmark  \\ 
\multicolumn{1}{l|}{{Cosentino19\cite{cosentino_search_2019}}}  & \xmark                     & \checkmark & FGSM           & \xmark &   \checkmark \\
\multicolumn{1}{l|}{{Li20\cite{li_towards_2020}}}  & \xmark                     & \checkmark & FGSM           & \xmark &   n.s.  \\
\multicolumn{1}{l|}{{Wang20\cite{wang_achieving_2020}}}  & \xmark                     & \checkmark & \xmark           & \xmark &   \checkmark  \\
\multicolumn{1}{l|}{{RST\cite{fu21_rst}}}  & \xmark                     & \checkmark & \xmark           & \xmark &   \checkmark  \\
\multicolumn{1}{l|}{{RobustBird\cite{chen_sparsity_2021}}}  & \checkmark                     & \checkmark & CW           & \xmark &   \checkmark\\ 
\multicolumn{1}{l|}{{AWT\cite{shi22_finding}}}  & \xmark                     & \checkmark & \xmark           & \xmark &   \checkmark \\

\multicolumn{1}{l|}{TwinRep\cite{li_twinrep_2023}}        & \checkmark                    & \checkmark & FGSM, BB     & \xmark &  \checkmark \\
\multicolumn{1}{l|}{BCS-P\cite{ozdenizci_training_2021}}          & \checkmark                    & \checkmark & FGSM BB      & \xmark &  \checkmark  \\
\multicolumn{1}{l|}{{DNR\cite{kundu_tunable_2020}}}      & \xmark                     & \checkmark & FGSM         & \xmark &  \checkmark   \\
\multicolumn{1}{l|}{{InTrain\cite{vemparala21_intrain}}}  & \xmark                     & \checkmark & CW           & \xmark &   \checkmark  \\ 
\multicolumn{1}{l|}{{FlyingBird\cite{chen_sparsity_2021}}}  & \checkmark                     & \checkmark & CW           & \xmark &   \checkmark   \\

\bottomrule
\end{tabular}%
}
\label{tab:ap_evals}
\end{table}

In~\autoref{tab:ap_evals}, we summarize the details of each AP adversarial evaluation and highlight their related issues. 
All in all, running adversarial attacks equals solving an optimization problem to find suitable perturbations that cause samples to be misclassified.
Such as any optimization problem, an inaccurate hyperparameter optimization or setup (including the implementation) can lead to sub-optimal solutions, thus failing to provide a better estimate for the problem, as well known in the adversarial ML literature~\cite{carlini19_onevaluating,pintor22_indicators,mura24_hofmn}. 
It is, therefore, fundamental to assess the validity of the adversarial evaluations carried out in AP papers.

\myparagraph{AA, PGD and Other attacks.} In columns ``AA", ``PGD", and ``Others", we indicate whether the AP method evaluated the robustness using, respectively, the AutoAttack framework~\cite{croce20_autoattack}, PGD~\cite{madry18-iclr} or/and any other attack. Through these columns, we directly highlight the diversity in the APs evaluations. 
AA, being an ensemble of four attacks, can be considered to supersede PGD; indeed, using multiple attacks at once can help ward off potential optimization issues, which are more likely using a single attack or few nearly identical versions~\cite{carlini19_onevaluating}. 
In addition, AA comprises a version of PGD (named AutoPGD) that automatically optimizes the step size and uses two loss versions, typically outperforming standard PGD~\cite{madry18-iclr}.  
We label with AA every AP method tested with AutoAttack.
We find a total of 9/26 AP methods tested using AA, while the remaining 17/26 use PGD as the predominant attack.\footnote{Note, however, that only a few AP methods were published before AA.}

\myparagraph{Robustness curve.} When using maximum-confidence attacks (e.g., PGD), the threat model encompasses a single perturbation budget $\epsilon$ (e.g., $\sfrac{8}{255}$ with \ellinf norm on the CIFAR10 dataset), thus returning the adversarial robustness of the model for that specific budget. 
However, evaluating adversarial robustness at a single perturbation size limits the evaluation, as it is not clear if robustness decreases more or less gracefully when increasing the perturbation size. 
Instead, a complete evaluation can be achieved through robustness curves, which plot the adversarial robustness against the perturbation norm~\cite{biggio18_wild,carlini19_onevaluating, pintor2021fast}. Using attacks such as PGD or AA, however, would require a substantial number of attack runs to plot such a curve, since a single attack run is bound to a fixed attack budget (see~\autoref{sect:adversarial}). In lieu of this demanding solution, minimum-norm attacks allow to straightforwardly plot the curve with a single attack run by returning adversarial examples not bounded to a single perturbation budget, but rather the smallest perturbation necessary~\cite{pintor2021fast}. 
Among the AP methods, although few also use minimum-norm attacks, none display robustness curves but rather clip the evaluation to a standard perturbation, with the unique exception of HYDRA, which plots the robustness over few discrete perturbation values with PGD attack~\cite{sehwag20-hydra}, yet not using a minimum norm to display a full curve. 

\myparagraph{Iter$>$10.} When using single attacks such as PGD, it is crucial to carefully select the hyperparameters to ensure convergence~\cite{carlini19_onevaluating}. Being PGD an iterative approach, although time-requiring, running the attack with a meaningful number of iterations improves the attack reliability~\cite{pintor22_indicators}. We thus list in ``Iter$>$10" the attacks using at least 10 iterations, which is a common default bare-minimum hyperparameter for the PGD attack, often leading to sub-optimal solutions. We find 3/26 setups using less than 10 iterations. 

\subsection{Benchmark Experimental Setting}\label{sect:exp_settings} 
Given the diversity in the experimental setups used to evaluate APs, and the often not top-performing adversarial evaluations, providing a benchmark for creating a uniform and fair evaluation of the pruned models becomes fundamental. 
In this section, we present the choices for the datasets, models, and sparsities adopted in our novel benchmark, thus laying the foundation for a comparable evaluation.  
In addition, we present the adversarial threat model employed for the benchmark to fairly and accurately re-evaluate the robustness of the pruned models.

\myparagraph{Datasets and architectures.} We mainly focus on the CIFAR10 and SVHN datasets on two specific architectures, ResNet18 and VGG16. We found combinations between these networks and datasets to be the best compromise with the available implementations and the most common ones in the analyzed papers.

\myparagraph{Sparsity rate.} Given the difference in the effect of structured (\structured) vs unstructured (\unstructured) pruning, we selected two different sets of sparsity rates, considering the lower tolerance to high sparsities of structured implementations (which, therefore, implies choosing a lower range of \sparsrate values). Thus, when using \structured, we prune each model with 50\%, 75\%, and 90\& \sparsrate, while when using \unstructured, we prune each model with 90\%, 95\%, and 99\% \sparsrate. 

\myparagraph{Attacks.} Following the discussion concerning the AP evaluations of~\autoref{tab:ap_evals}, to re-evaluate the robustness of the AP methods on the fixed models, datasets, and sparsity rates, we used the AutoAttack (AA) ensemble~\cite{croce20_autoattack}.
We run our evaluation on the entire test set, restricting to the \ellinf norm threat model, and using $\sfrac{8}{255}$ as perturbation budget $\epsilon$. 
To avoid having a single scalar robustness evaluation, we additionally used the Fast Minimum-Norm (FMN) attack~\cite{pintor2021fast}, through which we can collect multiple perturbation norms not confined to a single value and thus draw a complete robustness curve.
In addition, to have a more reliable estimate of the robustness, we optimized the hyperparameters of FMN for each model under test following the HO-FMN procedure~\cite{mura24_hofmn}.
We show how HO-FMN allows us to plot robustness curves, and report an analysis of such curves on the CIFAR10 dataset and \unstructured pruning, while more curves can be found in the available benchmark.

\subsection{Contributing to the Benchmark}
We welcome the submission of any new AP method in our publicly available benchmark and leaderboard. 
Contributing is simple and requires just three steps. 
\begin{itemize}
    \item Describing the AP method pipeline and specifics in the dedicated section, which will generate a JSON file.  
    \item Evaluating the checkpoints using our repository. This will compute and give the results for the given AP's checkpoints.
    \item Once the evaluation results and JSON data are received, authors are required to create a new issue in our repository using the dedicated template. Authors are required to add both the JSON entry and the evaluation results. 
\end{itemize}(i) 
We will then evaluate the submission and update our leaderboard.
Through our benchmark, we can evaluate the robustness curves of different AP methods for different sparsities and test every available checkpoint.
Finally, in addition to novel AP submissions, we welcome existing checkpoints (from the AP authors whose results have been reproduced in this paper) when deemed not up to par with their AP potential.

\subsection{Re-evaluation Results}\label{sect:results}
In the previous sections, we presented the benchmark details, which aim to act as a blueprint for evaluating AP methods.
In this section, we show the results obtained from re-evaluating the available implementations of the AP methods. 
We subdivide the results into four tables based on datasets and pruning structure. 
Then, given our taxonomy, we analyze and discuss the effect of each AP.
\begin{table*}[t]
\caption{The re-evaluation results for US pruning methods on CIFAR10. We report the clean and AutoAttack~\cite{croce20_autoattack} (AA) accuracies for the three US benchmark sparsity rates \sparsrate. We put in bold the top three APs for each sparsity/model combination.
}
\setlength{\tabcolsep}{17pt}
\centering
\resizebox{\textwidth}{!}{%
\begin{tabular}{@{}ccccccc@{}}
\toprule
 &
  \multicolumn{6}{c}{\textbf{Unstructured Pruning CIFAR10 (clean/AA)}} \\ \cmidrule(l){2-7} 
 &
  \multicolumn{3}{c|}{\textbf{ResNet18}} &
  \multicolumn{3}{c}{\textbf{VGG16}} \\
  
\multirow{-3}{*}{\textbf{Name}} &
  90 &
  95 &
  \multicolumn{1}{c|}{99} &
  90 &
  95 &
  99 \\ \midrule
{RADMM~\cite{ye19-radmm}} &
  80.54/43.68 &
  \textbf{79.33/42.56} &
  \multicolumn{1}{c|}{71.17/37.21} &
  74.76/39.92 &
  72.67/38.44 &
  57.69/31.30 \\
  
{HYDRA~\cite{sehwag20-hydra}} &
  76.74/43.34 &
  76.16/42.45 &
  \multicolumn{1}{c|}{\textbf{72.21/38.80}} &
  \textbf{78.31/43.81} &
  \textbf{76.58/42.61} &
  70.59/35.56 \\
{HARP~\cite{zhao23-harp}} &
  \textbf{83.38/45.40} &
  \textbf{83.38/45.69} &
  \multicolumn{1}{c|}{\textbf{83.11/45.50}} &
  \textbf{80.70/42.83} &
  \textbf{80.26/41.21} &
  \textbf{79.42/42.02} \\
  
{PwoA~\cite{tong22_pwoa}} &
  \textbf{83.29/45.35} &
  82.58/41.25 &
  \multicolumn{1}{c|}{76.33/28.95} &
  67.50/30.49 &
  65.85/26.39 &
  58.36/15.43 \\
{MAD~\cite{lee_masking_2022}} &
  73.67/41.10 &
  70.70/38.96 &
  \multicolumn{1}{c|}{58.90/29.26} &
  72.09/39.80 &
  70.45/38.10 &
  43.35/25.90 \\
    
{Li20~\cite{li_towards_2020}} &
  77.39/41.31 &
  73.54/39.29 &
  \multicolumn{1}{c|}{59.42/31.37} &
  75.66/39.26 &
  69.27/38.27 &
  58.49/31.24 \\
{RST~\cite{fu21_rst}} &
  60.92/14.31 &
  56.93/16.76 &
  \multicolumn{1}{c|}{48.90/15.16} &
  75.81/26.99 &
  71.45/23.94 &
  64.16/14.80 \\
  
{RobustBird~\cite{chen_sparsity_2021}} &
  78.16/43.35 &
  79.27/44.60 &
  \multicolumn{1}{c|}{69.36/37.08} &
  73.95/41.62 &
  76.16/41.80 &
  67.94/37.46 \\
{TwinRep~\cite{li_twinrep_2023}} &
  76.37/42.93 &
  73.19/41.47 &
  \multicolumn{1}{c|}{64.97/36.10} &
  75.36/41.84 &
  74.16/40.81 &
  \textbf{69.95/38.49} \\
    
{FlyingBird~\cite{chen_sparsity_2021}} &
  \textbf{80.69/46.49} &
  \textbf{77.42/46.10} &
  \multicolumn{1}{c|}{\textbf{75.40/42.02}} &
  \textbf{76.72/43.95} &
  \textbf{75.22/44.47} &
  \textbf{72.49/40.49}\\ 
  \bottomrule
\end{tabular}%
}
\label{tab:eval_US_cifar}
\end{table*}

\begin{table*}[t]
\setlength{\tabcolsep}{15pt}
\caption{The re-evaluation results for S (filter) pruning methods on CIFAR10. We report the clean and AutoAttack~\cite{croce20_autoattack} (AA) accuracies for the three US benchmark sparsity rates \sparsrate. We put in bold the single top AP for each setting.}
\centering
\setlength{\tabcolsep}{17pt}
\resizebox{\textwidth}{!}{%
\begin{tabular}{@{}ccccccc@{}}
\toprule
 &
  \multicolumn{6}{c}{\textbf{Structured Pruning CIFAR10 (clean/AA)}} \\ \cmidrule(l){2-7} 
 &
  \multicolumn{3}{c|}{\textbf{ResNet18}} &
  \multicolumn{3}{c}{\textbf{VGG16}} \\
  
\multirow{-3}{*}{\textbf{Name}} &
  50 &
  75 &
  \multicolumn{1}{c|}{90} &
  50 &
  75 &
  90 \\ \midrule
{RADMM~\cite{ye19-radmm}} &
  79.27/42.68 &
  78.81/40.79 &
  \multicolumn{1}{c|}{70.53/37.30} &
  74.58/39.67 &
  70.51/37.74 &
  58.58/31.79 \\
  
{HARP~\cite{zhao23-harp}} &
  77.38/42.73 &
  80.06/42.09 &
  \multicolumn{1}{c|}{77.88/41.59} &
  76.70/40.01 &
  73.61/39.14 &
  66.45/35.62 \\
{PwoA~\cite{tong22_pwoa}} &
  83.44/44.79 &
  81.77/37.85 &
  \multicolumn{1}{c|}{76.41/28.56} &
  66.33/30.15 &
  63.36/24.91 &
  57.71/18.39\\
  
{TwinRep~\cite{li_twinrep_2023}} &
  \textbf{79.90/45.58} &
  \textbf{79.37/45.21} &
  \multicolumn{1}{c|}{\textbf{78.41/44.30}} &
  \textbf{77.65/43.13} &
  \textbf{77.58/42.77} &
  \textbf{76.26/42.14}\\ 
  \bottomrule
\end{tabular}%
}
\label{tab:eval_S_cifar}
\end{table*} 

\begin{table*}[t]
\setlength{\tabcolsep}{15pt}
\caption{The re-evaluation results for US pruning methods on the SVHN dataset. We report the clean and AutoAttack~\cite{croce20_autoattack} (AA) accuracies for the three US benchmark sparsity rates \sparsrate. We put in bold the top three APs for each sparsity/model combination. 
}
\setlength{\tabcolsep}{17pt}
\centering
\resizebox{\textwidth}{!}{%
\begin{tabular}{@{}ccccccc@{}}
\toprule
 &
  \multicolumn{6}{c}{\textbf{Unstructured Pruning SVHN (CA/RA)}} \\ \cmidrule(l){2-7} 
 &
  \multicolumn{3}{c|}{\textbf{ResNet18}} &
  \multicolumn{3}{c}{\textbf{VGG16}} \\
  
\multirow{-3}{*}{\textbf{Name}} &
  90 &
  95 &
  \multicolumn{1}{c|}{99} &
  90 &
  95 &
  99 \\ \midrule
{RADMM~\cite{ye19-radmm}} &
  - &
  - &
  \multicolumn{1}{c|}{-} &
  62.25/44.40 &
  52.24/42.99 &
  \textbf{64.91/37.91} \\
  
{HYDRA~\cite{sehwag20-hydra}} &
  90.95/44.12 &
  89.91/45.29 &
  \multicolumn{1}{c|}{85.71/34.20} &
  \textbf{87.89/45.85} &
  \textbf{87.95/44.57} &
  80.85/40.30 \\
{HARP~\cite{zhao23-harp}} &
  \textbf{92.96/45.39} &
  92.75/45.95 &
  \multicolumn{1}{c|}{93.38/34.42} &
  92.69/44.00 &
  92.25/44.17 &
  \textbf{90.60/44.36} \\
  
{PwoA~\cite{tong22_pwoa}} &
  92.41/42.66 &
  92.21/39.50 &
  \multicolumn{1}{c|}{90.05/29.58} &
  89.33/38.95 &
  89.08/35.20 &
  84.47/21.46 \\
{MAD~\cite{lee_masking_2022}} &
  - & 
  - &
  \multicolumn{1}{c|}{-} &
  89.42/37.46 & 
  86.40/24.90 &
  - \\
 
{Li20~\cite{li_towards_2020}} &
  89.95/43.62 &
  55.04/19.98 &
  \multicolumn{1}{c|}{36.71/13.09} &
  53.69/26.31 &
  48.24/20.39 &
  45.88/14.56 \\
{RST~\cite{fu21_rst}} &
  79.89/34.15 &
  74.90/31.94 &
  \multicolumn{1}{c|}{61.55/25.35} &
  88.74/43.99 &
  87.64/41.91 &
  88.42/41.25 \\
  
{RobustBird~\cite{chen_sparsity_2021}} &
  \textbf{91.00/46.23} &
  \textbf{90.18/47.26} &
  \multicolumn{1}{c|}{86.12/42.62} &
  89.04/42.81 &
  88.24/41.64 &
  - \\
{TwinRep~\cite{li_twinrep_2023}} &
  \textbf{88.90/46.72} &
  \textbf{88.59/47.16} &
  \multicolumn{1}{c|}{\textbf{85.09/43.44}} &
  \textbf{87.22/45.54} &
  \textbf{89.70/44.33} &
  \textbf{86.03/43.55} \\
    
{FlyingBird~\cite{chen_sparsity_2021}} &
  92.60/39.81 &
  \textbf{91.14/47.43} &
  \multicolumn{1}{c|}{92.15/41.80} &
  \textbf{91.05/49.04} &
  \textbf{91.12/49.94} &
  \textbf{90.03/48.80}\\ 
  \bottomrule
\end{tabular}%
}
\label{tab:eval_US_svhn}
\end{table*}

\begin{table*}[t]
\setlength{\tabcolsep}{17pt}
\caption{The re-evaluation results for S pruning methods on SVHN. In bold, the top method for each combination.}
\centering
\resizebox{\textwidth}{!}{%
\begin{tabular}{@{}ccccccc@{}}
\toprule
 &
  \multicolumn{6}{c}{\textbf{Structured Pruning SVHN (clean/AA)}} \\ \cmidrule(l){2-7} 
 &
  \multicolumn{3}{c|}{\textbf{ResNet18}} &
  \multicolumn{3}{c}{\textbf{VGG16}} \\
  
\multirow{-3}{*}{\textbf{Name}} &
  50 &
  75 &
  \multicolumn{1}{c|}{90} &
  50 &
  75 &
  90 \\ \midrule
{RADMM~\cite{ye19-radmm}}                   & - & - & \multicolumn{1}{c|}{-} & - & - & - \\  
{HARP~\cite{zhao23-harp}}                     & \textbf{91.72/45.82} & \textbf{92.07/46.80} & \multicolumn{1}{c|}{\textbf{91.03/45.25}} & 91.53/44.10 & 89.06/42.45 & 87.89/39.25 \\
{PwoA~\cite{tong22_pwoa}}                     & 92.56/41.68 & 92.61/38.69 & \multicolumn{1}{c|}{91.42/31.69} & 89.16/39.09 & 89.22/33.89 & 87.17/24.55 \\ 
{TwinRep~\cite{li_twinrep_2023}}                  & 90.71/37.33 & 88.71/45.28 & \multicolumn{1}{c|}{85.44/45.10} & \textbf{89.91/45.82} & \textbf{87.10/43.26} & \textbf{89.61/44.83}\\ 
  \bottomrule
\end{tabular}%
}
\label{tab:eval_S_svhn}
\end{table*}

\begin{table*}[t]
\setlength{\tabcolsep}{29pt}
\caption{The pretrained models, evaluated with AutoAttack, for each of the \Amethods. For PwoA~\cite{tong22_pwoa}, which distills on a robust pretrained model, we used the HARP~\cite{zhao23-harp} robust checkpoints.}
\centering
\resizebox{\textwidth}{!}{%
\begin{tabular}{ccccc}
\hline
\multirow{3}{*}{\textbf{Name}} & \multicolumn{4}{c}{\textbf{Pretrained (clean/AA)}}                                 \\ \cline{2-5} 
                               & \multicolumn{2}{c|}{\textbf{CIFAR10}}          & \multicolumn{2}{c}{\textbf{SVHN}} \\
                               & ResNet18    & \multicolumn{1}{c|}{VGG16}       & ResNet18        & VGG16           \\ \hline
{RADMM~\cite{ye19-radmm}}                   & 79.06/44.42 & \multicolumn{1}{c|}{74.65/41.59} & 84.90/41.23     & 78.99/43.89     \\
{HYDRA~\cite{sehwag20-hydra}}                    & 79.28/46.92 & \multicolumn{1}{c|}{78.12/42.78} & 90.50/44.67     &   88.60/45.66   \\
{HARP~\cite{zhao23-harp}}                     & 81.30/49.48 & \multicolumn{1}{c|}{80.18/45.09} & 90.70/42.08     & 88.66/44.62     \\
{MAD~\cite{lee_masking_2022}}                      & 80.27/43.50 & \multicolumn{1}{c|}{76.06/40.30} & 88.13/43.77     & 87.56/45.39     \\
{PwoA~\cite{tong22_pwoa}}                     & 81.30/49.48 & \multicolumn{1}{c|}{80.18/45.09} & 90.70/42.08     & 88.66/44.62\\ 
  \bottomrule   
\end{tabular}%
}
\label{tab:pretrained}
\end{table*}

\myparagraph{Available implementations.} 
We found a total of 11 available AP implementations, which we subdivided based on structured (\structured) and unstructured (\unstructured) pruning. 
Among these, we total 4 \structured and 10 \unstructured available pruning implementations, thus pruning 14 models for each of the 12 dataset/network/sparsity combinations for a total of 168 models. 
We point out that, although two further implementations were available, we could not solve their bugs or extend to the selected networks~\cite{ozdenizci_training_2021,zhong2023adv_robust_gkp}. 
Among the available ones, we specify that we experienced multiple issues in: (i) reproducing the results for MAD~\cite{lee_masking_2022} and RST~\cite{fu21_rst} on CIFAR10 with orders of 5 and 15 percentage points, respectively, and for which we welcome checkpoints used for the paper results; (ii) fitting the model for MAD~\cite{lee_masking_2022} on SVHN, for which we had to modify the training procedures from the original one but failed in many occasions (hence the ``-" entries in~\autoref{tab:eval_US_svhn}).
In addition, SVHN was not originally implemented in multiple of the available APs; we thus extended the code in RADMM~\cite{ye19-radmm} (for which we occasionally had troubles in fitting the models), PwoA~\cite{tong22_pwoa}, RobustBird~\cite{chen_sparsity_2021}, FlyingBird~\cite{chen_sparsity_2021} and Li20~\cite{li_towards_2020}.
In turn, we will mainly derive our conclusions and analysis from the CIFAR10 results and then only validate on SVHN.
To conclude, we specify that for HYDRA~\cite{sehwag20-hydra}, leveraging additional data following~\cite{carmon19-unlabeled}, we used the standard dataset training to provide a fair comparison.

\myparagraph{Pruning structure.} 
On both CIFAR10 and SVHN, we notice from the results of~\autoref{tab:eval_US_cifar} and~\autoref{tab:eval_S_cifar}, how \unstructured pruning is less sensitive to higher sparsities than \structured. 
On the shared sparsity of 90\%, except for TwinRep~\cite{li_twinrep_2023} on CIFAR10, all the AP methods hold a constantly greater accuracy and robustness against their structured counterpart.
Pruning single weights as opposed to entire structures, as widely known in the field~\cite{liu2023lessons}, gives indeed higher flexibility and results in likewise higher performance.

\myparagraph{Optimizing layer-wise sparsity matters.}
In~\autoref{tab:eval_US_cifar} for \unstructured on CIFAR10, where we put in bold the results from the top-3 APs, we notice how FlyingBird~\cite{chen_sparsity_2021} and HARP~\cite{zhao23-harp} consistently outperform the other AP methods and represent the only two methods reaching over 40\% robust accuracy at 99\% sparsities on CIFAR10. 
We suppose their constant advantage stems from optimizing the layer-wise sparsity. 
In fact, both FlyingBird and HARP besides allowing different sparsities within each layer, additionally find an optimal strategy: \ie, they find an optimal sparsity for each layer of the network, ultimately satisfying the desired global \sparsrate. 
Also, Heracles~\cite{zhao22_heracles}, on top of which HARP is built, demonstrated the efficacy of optimizing the \sparsrate in each layer by improving both RADMM and HYDRA through an optimal layer-wise sparsity. 
Similarly, also TwinRep is based on a simpler (not optimized) \gglobal pruning locality and is often found to be one of the top performing AP methods in \structured pruning.
Just like the flexibility of \unstructured pruning allows attaining higher performances than \structured, we suppose that also the flexibility given by global pruning, enhanced by optimizing the layer-wise sparsity, enables higher robustness and accuracy.

\myparagraph{Is complexity rewarding?}
The analyzed AP methods are typically associated with complex pipelines and criterion designs. 
While comparing to a simpler pipeline is often demanding and, in general, not necessary, comparing to a simpler criterion is simpler and helps understand the true benefit of designing a complex and articulated pruning criterion. 
Therefore, considering LWM as a naive criterion for \unstructured and filter L1 norm for \structured~\cite{han15-learning, haocvpr16-filters}, we questioned how frequently papers adopting LIS, RELWM, or SOLWM criteria compared with a naive one.  
In general, we found out to be quite rare for the surveyed AP methods to compare to naive criteria. 
We believe that such comparison helps validate the complexity of the AP and, most importantly, helps understand how much adopting such complex and often time-requiring criterion pays off.

\begin{table}[ht]
\centering
\caption{Robustness of LWM and LWM+LAMP \gglobal strategy on CIFAR10 models. }
\label{tab:lwm_lamp}
\setlength{\tabcolsep}{10pt}
\resizebox{0.44\textwidth}{!}{%
\begin{tabular}{@{}ccccc@{}}
\toprule
\multirow{3}{*}{\textbf{Sparsity}} & \multicolumn{2}{c}{\textbf{ResNet18}} & \multicolumn{2}{c}{\textbf{VGG16}} \\ \cmidrule(l){2-5} 
                                   & \textbf{LWM}      & \textbf{LAMP}     & \textbf{LWM}    & \textbf{LAMP}    \\ \midrule 
90\%                               & 39.69             & 42.19             & 35.54           & 40.12            \\
95\%                               & 37.57             & 39.94             & 32.47           & 37.81                 \\ 
99\%                               & 30.04             & 36.98             & 27.52                & 33.07            \\
\bottomrule
\end{tabular}
}
\end{table}
\myparagraph{Flexible and cheap solution.}
We have shown how AP methods typically reach higher adversarial robustness when they use \gglobal pruning, which is even higher when the \sparsrate in each layer is optimized. 
In addition, we have seen that most of the papers presenting AP methods do not compare their often complex criteria to a naive one, such as LWM. 
We thus question how much gain can be obtained by a naive and cheap criterion such as LWM when combined with a layer-wise strategy (\ie, different sparsity in each layer), and whether such gain could also be capable of surpassing more complex criterion using \local pruning. 
Therefore, we combined LWM with a different \sparsrate strategy for each layer. 
However, instead of optimizing the layerwise \sparsrate such as in HARP (which would clearly increase performances but also complexity), we make use of cheaper approaches such as Layer-wise Adaptive Magnitude Pruning LAMP~\cite{lee21_iclr}, which computes an ideal layerwise sparsity without any additional cost, thus preserving the low-cost of a naive criterion but improving the performances of LWM.  
Interestingly, through our results in~\autoref{tab:lwm_lamp}, we show that LWM has a significant gain from using the \gglobal strategy of LAMP, to the extent that some of the more complex and time-requiring APs retain lower robustness. 
This corroborates results from prior work (HARP and Heracles~\cite{zhao22_heracles, zhao23-harp}) showing the benefits of non-uniformity on their corresponding competing AP methods~\cite{zhao22_heracles, zhao23-harp}. 
Most importantly, it shows that complex AP methods can often be surpassed by simple and low complexity solutions increasing the ``flexibility" of the chosen weights to be pruned.

\myparagraph{Unreliable evaluations.}
While many of the published AP methods do not test robustness with AA, we find only two methods among the available implementations never testing any model on AA, RADMM~\cite{ye19-radmm} and Li20~\cite{li_towards_2020}.
However, for the rest of the available APs testing on AA, we notice how not every paper extensively tests all model-sparsity combinations on AA, but rather executes just a few trials on selected pairs of models while only relying on PGD in an extensive way.
Therefore, by extending the evaluations using AA, it is possible to show how APs have often overestimated robustness.
Still, it is hard to exactly compare the results of our re-evaluation with the original papers' results, since models in their papers are pruned to sparsities that do not necessarily match our benchmark setting. 
We specify that our benchmark works right toward this direction: providing a template for previous and new APs to be fairly and reliably tested and compared.

Despite these limitations, we report results from the original AP papers for similar (or the same, when possible) sparsities on CIFAR10, aiming to quantify the extent to which robustness has been overestimated in available APs. 
\begin{table*}[t]
\caption{Summary of robustness overestimation in AP papers for CIFAR10 for both \unstructured and \structured pruning. We consider, for each AP, the model with \sparsrate matching ours, or when not available, the closest and lower of our sparsities (\ie, $93.75$ is compared to our $90$).
We indicate the paper's PGD robustness with ``Orig." and with ``Ours," the one computed with AA, and the difference between the two.}
\centering
\setlength{\tabcolsep}{10pt}
\renewcommand{\arraystretch}{1.2}
\resizebox{0.95\textwidth}{!}{%
\begin{tabular}{clllllllllll}
\toprule
 &
  \multicolumn{5}{c}{\textbf{ResNet18}} &
   &
  \multicolumn{5}{c}{\textbf{VGG16}} \\ \cline{1-6} \cline{8-12} 
 &
  \textbf{Name} &
  \textbf{\sparsrate} &
  \textbf{Orig.} &
  \textbf{Ours} &
  \textbf{Diff.} &
   &
  \textbf{Name} &
  \textbf{\sparsrate} &
  \textbf{Original} &
  \textbf{Ours} &
  \textbf{Diff.} \\ \midrule
\multirow{7}{*}{US} &  
  RADMM~\cite{ye19-radmm} &
  \multicolumn{1}{l|}{93.75} &
  47.00 &
  43.68 &
  \textbf{3.32} &
   &
  RADMM~\cite{ye19-radmm} &
  \multicolumn{1}{l|}{93.75} &
  45.00 &
  39.92 &
  \textbf{5.08} \\ \cline{2-12} 
 &
  \multirow{2}{*}{RobustBird~\cite{chen_sparsity_2021}} &
  \multicolumn{1}{l|}{90.00} &
  49.09 &
  43.35 &
  \textbf{5.74} &
   &
  \multirow{2}{*}{RobustBird~\cite{chen_sparsity_2021}} &
  \multicolumn{1}{l|}{\multirow{2}{*}{90.00}} &
  \multirow{2}{*}{47.09} &
  \multirow{2}{*}{41.62} &
  \multirow{2}{*}{\textbf{5.74}} \\
 &
   &
  \multicolumn{1}{l|}{95.00} &
  47.53 &
  44.60 &
  \textbf{2.93} &
   &
   &
  \multicolumn{1}{l|}{} &
   &
   &
   \\ \cline{2-12} 
 &
  \multirow{2}{*}{TwinRep~\cite{li_twinrep_2023}} &
  \multicolumn{1}{l|}{90.00} &
  49.30 &
  42.93 &
  \textbf{6.37} &
   &
  \multirow{2}{*}{TwinRep~\cite{li_twinrep_2023}} &
  \multicolumn{1}{l|}{\multirow{2}{*}{-}} &
  \multirow{2}{*}{-} &
  \multirow{2}{*}{-} &
  \multirow{2}{*}{\textbf{-}} \\
 &
   &
  \multicolumn{1}{l|}{95.00} &
  47.10 &
  41.47 &
  \textbf{5.63} &
   &
   &
  \multicolumn{1}{l|}{} &
   &
   &
   \\ \cline{2-12} 
 &
  \multirow{2}{*}{FlyingBird~\cite{chen_sparsity_2021}} &
  \multicolumn{1}{l|}{90.00} &
  50.97 &
  46.49 &
  \textbf{4.48} &
   &
  \multirow{2}{*}{FlyingBird~\cite{chen_sparsity_2021}} &
  \multicolumn{1}{l|}{\multirow{2}{*}{90.00}} &
  \multirow{2}{*}{48.45} &
  \multirow{2}{*}{43.95} &
  \multirow{2}{*}{\textbf{4.50}} \\
 &
   &
  \multicolumn{1}{l|}{95.00} &
  49.62 &
  46.10 &
  \textbf{3.32} &
   &
   &
  \multicolumn{1}{l|}{} &
   &
   &
   \\ \midrule 
\multirow{4}{*}{S} &
  \multirow{3}{*}{RADMM~\cite{ye19-radmm}} &
  \multicolumn{1}{l|}{50.00} &
  45.00 &
  42.68 &
  \textbf{2.32} &
   &
  \multirow{3}{*}{RADMM~\cite{ye19-radmm}} &
  \multicolumn{1}{l|}{50.00} &
  42.00 &
  39.67 &
  \textbf{2.32} \\
 &
   &
  \multicolumn{1}{l|}{75.00} &
  44.00 &
  40.79 &
  \textbf{3.21} &
   &
   &
  \multicolumn{1}{l|}{75.00} &
  41.00 &
  37.74 &
  \textbf{3.26} \\
 &
   &
  \multicolumn{1}{l|}{93.75} &
  39.00 &
  37.30 &
  \textbf{1.70} &
   &
   &
  \multicolumn{1}{l|}{93.75} &
  35.00 &
  31.79 &
  \textbf{3.21} \\ \cline{2-12} 
 & 
  TwinRep~\cite{li_twinrep_2023} &
  \multicolumn{1}{l|}{50.00} &
  48.60 &
  45.58 &
  \textbf{3.02} &
   & 
  TwinRep~\cite{li_twinrep_2023} &
  \multicolumn{1}{l|}{-} &
  - &
  - &
  \textbf{-} \\ \bottomrule
\end{tabular}%
}
\label{tab:overestimate_cifar}
\end{table*}

In~\autoref{tab:overestimate_cifar}, we show the summary of the robustness overestimation in AP papers. We use results from original papers matching (or the closest to) our benchmark sparsities for which no AA~\cite{croce20_autoattack} evaluation has been reported. In fact, despite being only few papers among the ones with available implementation not evaluated with AA (see~\autoref{tab:ap_evals}), we find an actual complete and extensive AA evaluation to be not common, while it is much more frequent to find extensive PGD evaluations and few AA ones. Therefore, it is possible to select results evaluated only on PGD and compare them to our AA re-evaluation. In the case of RADMM~\cite{ye19-radmm}, where the reported values have \sparsrate 93.75\%, we compare to our evaluation at 90\%. Although the former is supposed to retain lower robustness, it actually holds a higher robustness estimate than our re-evaluation with AA on a smaller sparsity, thus indicating an evident overestimation. Overall, compared to our AA re-evaluation, we find multiple papers overestimating the robustness.

\myparagraph{Robustness curves.}
To have a comprehensive evaluation on multiple perturbation norms and thus plot robustness curves, we use one of the most recent minimum-norm attacks, known as HO-FMN~\cite{mura24_hofmn}, which optimizes the attack hyperparameters to find the best configuration on which to run the attack for the model under test. 
Using a minimum-norm attack, the robustness evaluation is not bounded to a single scalar value (e.g., $\sfrac{8}{255}$), since the goal is to find the smallest perturbation norm.
Therefore, as discussed in~\autoref{sect:adversarial}, the adversarial examples found by the attack are associated with multiple perturbations, which makes the evaluation of the curve straightforward and efficient~\cite{carlini19_onevaluating, pintor2021fast}. 

\begin{figure*}[ht]   
\centering
   \includegraphics[width=0.8\textwidth]{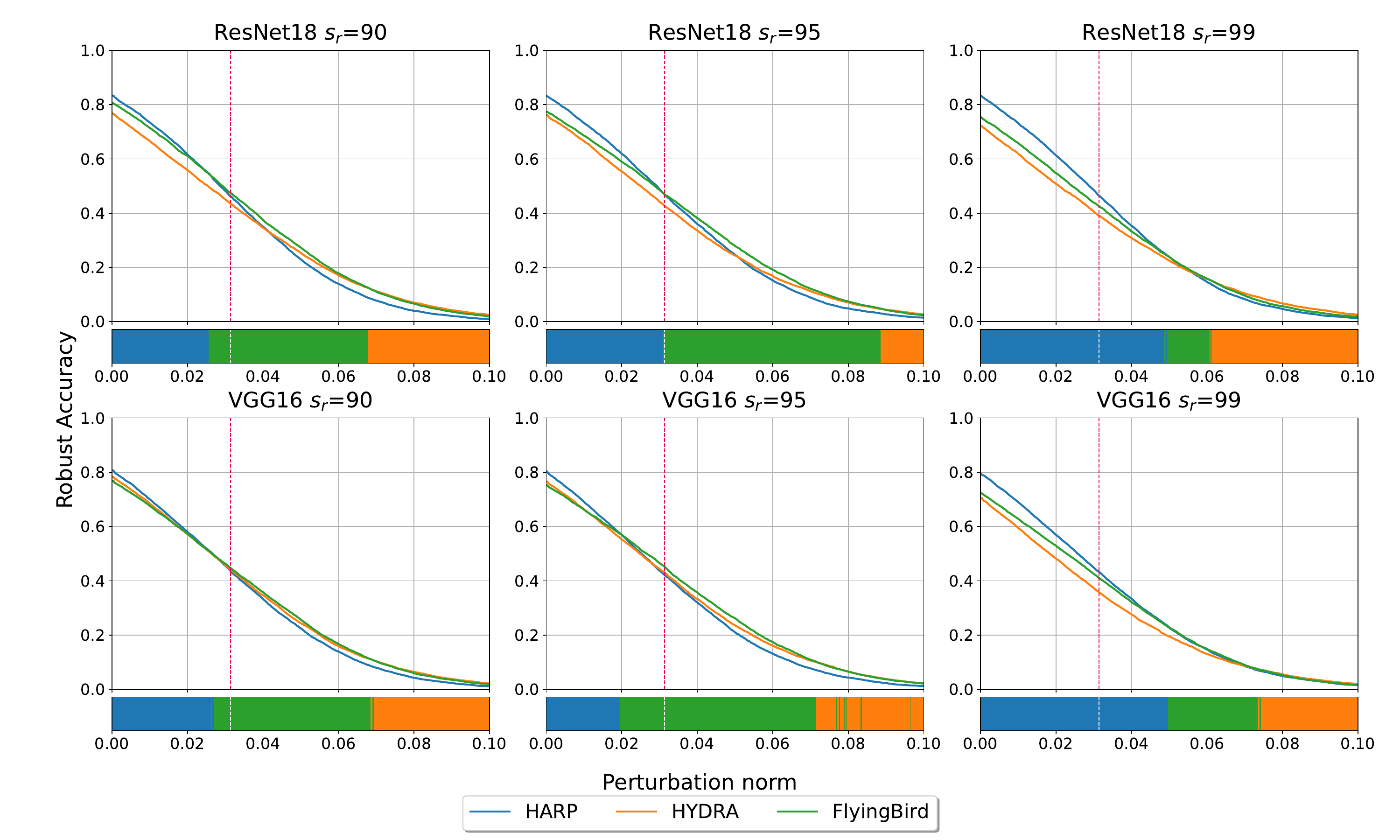}
\caption{The robustness curves of HARP, HYDRA, and FlyingBird for each sparsity rate with \unstructured pruning on CIFAR10. The plots have been created using the HO-FMN optimization procedure for the FMN attack~\cite{pintor2021fast,mura24_hofmn}. The curves show robust accuracy against the perturbation norm $\epsilon$, where the value $\sfrac{8}{255}$ is represented with a vertical dotted line. The bar subplot shows the color of the most robust model, describing how the ``leader model" varies over perturbation.}
\label{fig:sec_curve_top3_cifar10_us}
\end{figure*}

In reference to HO-FMN, we run the search for the attack hyperparameters on 2048 samples for each pruned model.
We then run the attack on the remaining samples using the best hyperparameters found on the DLR loss, SGD optimizer and Cosine Annealing scheduler.
We show, in~\autoref{fig:sec_curve_top3_cifar10_us}, the robustness curves for the top-3 APs on the CIFAR10 dataset, HYDRA, HARP, and FlyingBird, for each of the three sparsities of \unstructured. 
Robustness curves show the decrease of robustness as the perturbation grows, which does not necessarily reflect the robustness evaluation computed on a scalar perturbation (\eg, through AA). 
To highlight such an aspect, we show a bar subplot in~\autoref{fig:sec_curve_top3_cifar10_us} using the color of the most robust model, thus showing how, as the perturbation grows, the robustness across models can change significantly.
When the perturbation is small, we find HARP to be the most robust model. Instead, FlyingBird is typically the most robust around just a small window around $\sfrac{8}{255}$, which is the typical perturbation norm used to evaluate the models. 
As the perturbation grows, HYDRA is instead constantly the most robust model. 
Using robustness curves represents a thus fundamental tool to guarantee a complete adversarial robustness evaluation and shows how a single scalar evaluation might not always be sufficient. 

\myparagraph{Extension on ImageNet dataset.} 
We additionally expand on the ImageNet dataset and ResNet50 architecture for the HARP~\cite{zhao23-harp} and HYDRA~\cite{sehwag20-hydra} methods, resulting in a comparison of AP methods for a bigger-scale dataset.
As for the other datasets, we allow the online benchmark to host ImageNet models, thus providing a further comparison for AP methods. 
We show, in~\autoref{tab:eval_US_imagenet}, the results for \unstructured pruning, which, similar to the results for different datasets and architectures, confirm the benefits of the optimized layer-wise sparsity of HARP, compared to HYDRA.
\begin{table}[ht]
\centering
\setlength{\tabcolsep}{25pt}
\caption{The re-evaluation results for \unstructured pruning methods on Imagenet and ResNet50.}
\label{tab:eval_US_imagenet}
\resizebox{0.48\textwidth}{!}{%
\begin{tabular}{@{}ccc@{}}
\toprule
\multicolumn{1}{c}{\textbf{Sparsity}} & \multicolumn{1}{c}{\textbf{HARP}} & \multicolumn{1}{c}{\textbf{HYDRA}} \\ 
\midrule
90\%                               & \textbf{23.35}                     & 21.36            \\
95\%                               & \textbf{22.88}                     & 19.71                 \\ 
99\%                               & \textbf{12.56}                     & 10.37            \\  
\bottomrule
\end{tabular}
}
\end{table}

\section{Related Work}
\label{sect:related}
Adversarial pruning methods represent a set of fundamental techniques capable of producing robust pruned models. 
However, pruning techniques are associated with a possibly great diversity, in addition to a nontrivial design complexity. 
It is thus fundamental to taxonomize such methods, analyze their design, and evaluate them in a fair and accurate way to understand the effect of pipelines and specifics.
While a great number of AP methods have been presented, only a few works in the literature have attempted to analyze the relationship between pruning and adversarial robustness. 
In~\cite{jordao_effect_2021}, the authors analyze the effect of pruning on robustness, showing that pruning can help robustness without adversarial training, thus by acting as a regularizer: however, the pruning methods considered are limited to structured pruning, and their robustness is tested against a single-step FGSM attack, leading to potentially unreliable evaluations.
In~\cite{vora_benchmarking_2023}, although conversely adopting a proper selection of attacks to test the robustness of models, the considered pruning methods are limited to naive structured pruning methods. 
Slightly closer to our work, in~\cite{merkle_pruning_2021}, the authors introduce the definition of locality, structure and criterion and test only the adversarial robustness of naive pruning methods. 
Regarding APs instead, in~\cite{piras23_thinice}, the authors present a re-evaluation limited to the HYDRA and RADMM APs, mainly focusing on the effect of pruning on the dynamics of the models' decision boundary.
Then, in~\cite{pavlitska23_relationship}, the authors present a review of the experimental setup used in multiple compression methods for adversarial robustness.  

Focusing on clean accuracy instead, the survey and taxonomy in~\cite{cheng2023survey} employs a similar approach to ours by surveying multiple work and creating an overall taxonomy that yet differs in both entries and structure, as we focus specifically on adversarial pruning methods (which, instead, are not considered on the related work). 
Therefore, differently from existing literature, we focus on the isolated case of adversarial pruning methods, which require specific attention due to their robustness-oriented complex designs and, most importantly, to the care in which adversarial robustness evaluations need to be set up. 
Towards a better comprehension of such methods, indeed, our taxonomy is built to thoroughly classify the methods and understand, with our re-evaluation and proposed benchmark, the overall effect of each AP design in a comparable, fair, and accurate adversarial experimental setting.

\section{Conclusion and Future Work}  
\label{sect:conclusions}
In this work, we proposed a taxonomy of AP methods with the goal of comparing the current AP methods' designs.
In addition, we reviewed existing AP's adversarial evaluations and found multiple issues related to their reliability. 
Towards a fair, reliable comparison of AP methods, we thus proposed our benchmark, which allows AP methods to be evaluated in the same experimental and adversarial setup. 
Hence, we used our benchmark to re-evaluate the available implementation of AP methods, which we studied alongside the taxonomy to associate the different designs with their effects. 
With our work, we provide a unique framework to ``unwrap" and explain the design of these methods, while also allowing to test them uniformly and compare the results. The ultimate goal is thus to pave the way for future adversarial pruning research, which can now benefit from a reference point on which to compare and test their novel designs.

In conclusion, we acknowledge that our work is limited to neural networks, not expanding on more complex architectures such as Vision Transformers (ViT). However, unfortunately, work on ViT robustness and ViT pruning follow two different paths which have yet to meet. While existing work on pruning ViT~\cite{chen21_chasing} provide the foundation in fact, integrating these methods with adversarial robustness represents a non-trivial and yet unexplored challenge. In this regard, we believe that our taxonomy and benchmark can be leveraged by future work to investigate AP on transformers.
Additionally, different applications of adversarial pruning, such as in natural language processing, remain likewise rather unexplored. We thus hope that this work inspires future research to tackle these challenges and advance adversarial pruning toward broader applicability and impact.

\section*{Acknowledgements}
This work was carried out while Giorgio Piras was enrolled in the Italian National Doctorate on AI run by the Sapienza University of Rome in collaboration with the University of Cagliari. 
This work was partially supported by the NRRP MUR program funded by the EU-NGEU under the projects SERICS (PE00000014) and FAIR (PE00000013); by the European Union's Horizon Europe Research and Innovation Programme under the project Sec4AI4Sec (grant agreement no. 101120393) and ELSA (grant agreement no. 101070617); and by Fondazione di Sardegna under the project ``TrustML: Towards Machine Learning that Humans Can Trust’’, CUP: F73C22001320007.

\bibliography{main2}

\begin{thebibliography}{10}
\expandafter\ifx\csname url\endcsname\relax
  \def\url#1{\texttt{#1}}\fi
\expandafter\ifx\csname urlprefix\endcsname\relax\def\urlprefix{URL }\fi
\expandafter\ifx\csname href\endcsname\relax
  \def\href#1#2{#2} \def\path#1{#1}\fi

\bibitem{belking19-reconciling}
M.~Belkin, D.~Hsu, S.~Ma, S.~Mandal, Reconciling modern machine-learning
  practice and the classical bias{\textendash}variance trade-off, Proceedings
  of the National Academy of Sciences.

\bibitem{frankle_lth19}
J.~Frankle, M.~Carbin, The lottery ticket hypothesis: Finding sparse, trainable
  neural networks, in: ICLR, 2019, pp. 1--10.

\bibitem{han15-learning}
S.~Han, J.~Pool, J.~Tran, W.~J. Dally, Learning both weights and connections
  for efficient neural networks, in: NeurIPS, 2015, pp. 1--10.

\bibitem{vanhoucke-nips11}
V.~Vanhoucke, A.~Senior, M.~Z. Mao, Improving the speed of neural networks on
  cpus, in: Deep Learning and Unsupervised Feature Learning Workshop, NIPS,
  2011, pp. 1--10.

\bibitem{hinton15-distillation}
G.~Hinton, O.~Vinyals, J.~Dean, Distilling the knowledge in a neural network,
  in: NIPS Deep Learning and Representation Learning Workshop, 2015, pp. 1--10.

\bibitem{lecun89-obd}
Y.~LeCun, J.~S. Denker, S.~A. Solla, Optimal brain damage, in: NIPS, 1989, pp.
  1--10.

\bibitem{molchanov16-pruning}
P.~Molchanov, S.~Tyree, T.~Karras, T.~Aila, J.~Kautz, Pruning convolutional
  neural networks for resource efficient inference, in: ICLR, 2017, pp. 1--10.

\bibitem{han16-deepcompression}
S.~Han, H.~Mao, W.~J. Dally, Deep compression: Compressing deep neural network
  with pruning, trained quantization and huffman coding, in: ICLR, 2016, pp.
  1--10.

\bibitem{He_2017_ICCV}
Y.~He, X.~Zhang, J.~Sun, Channel pruning for accelerating very deep neural
  networks, in: ICCV, 2017, pp. 1--10.

\bibitem{biggio13-ecml}
B.~Biggio, I.~Corona, D.~Maiorca, B.~Nelson, N.~\v{S}rndi\'{c}, P.~Laskov,
  G.~Giacinto, F.~Roli, Evasion attacks against machine learning at test time,
  in: ECML-PKDD, 2013.

\bibitem{szegedy_intriguing_2014}
C.~Szegedy, W.~Zaremba, I.~Sutskever, J.~Bruna, D.~Erhan, I.~J. Goodfellow,
  R.~Fergus, Intriguing properties of neural networks, in: ICLR, 2014.

\bibitem{madry18-iclr}
A.~{Madry}, A.~{Makelov}, L.~{Schmidt}, D.~{Tsipras}, A.~{Vladu}, Towards deep
  learning models resistant to adversarial attacks, in: ICLR, 2018, pp. 1--10.

\bibitem{croce20_autoattack}
F.~Croce, M.~Hein, Reliable evaluation of adversarial robustness with an
  ensemble of diverse parameter-free attacks, in: ICML, 2020, pp. 1--10.

\bibitem{zhang19-trades}
H.~Zhang, Y.~Yu, J.~Jiao, E.~P. Xing, L.~E. Ghaoui, M.~I. Jordan, Theoretically
  principled trade-off between robustness and accuracy, in: ICML, 2019, pp.
  1--10.

\bibitem{biggio18_wild}
B.~Biggio, F.~Roli, Wild patterns: Ten years after the rise of adversarial
  machine learning, Pattern Recognition 84 (2018) 317--331.

\bibitem{sehwag20-hydra}
V.~Sehwag, S.~Wang, P.~Mittal, S.~Jana, {HYDRA:} pruning adversarially robust
  neural networks, in: NeurIPS, 2020, pp. 1--10.

\bibitem{ye19-radmm}
S.~Ye, X.~Lin, K.~Xu, S.~Liu, H.~Cheng, J.~Lambrechts, H.~Zhang, A.~Zhou,
  K.~Ma, Y.~Wang, Adversarial robustness vs. model compression, or both?, in:
  ICCV, 2019, pp. 1--10.
\newblock \href {http://dx.doi.org/10.1109/ICCV.2019.00020}
  {\path{doi:10.1109/ICCV.2019.00020}}.

\bibitem{vemparala21_intrain}
M.-R. Vemparala, N.~Fasfous, A.~Frickenstein, S.~Sarkar, Q.~Zhao, S.~Kuhn,
  L.~Frickenstein, A.~Singh, C.~Unger, N.-S. Nagaraja, C.~Wressnegger,
  W.~Stechele, Adversarial {Robust} {Model} {Compression} using {In}-{Train}
  {Pruning}, in: CVPRW, 2021, pp. 1--10.

\bibitem{cosentino_search_2019}
J.~Cosentino, F.~Zaiter, D.~Pei, J.~Zhu, The {Search} for {Sparse}, {Robust}
  {Neural} {Networks}, number: arXiv:1912.02386 arXiv:1912.02386 [cs, stat]
  (Dec. 2019).

\bibitem{zhao23-harp}
Q.~Zhao, C.~Wressnegger, Holistic adversarially robust pruning, in: ICLR, 2023,
  pp. 1--10.

\bibitem{tong22_pwoa}
T.~Jian, Z.~Wang, Y.~Wang, J.~G. Dy, S.~Ioannidis, Pruning adversarially robust
  neural networks without adversarial examples, in: ICDM, 2022, pp. 1--10.

\bibitem{sandler18_mobilenetv2}
M.~Sandler, A.~Howard, M.~Zhu, A.~Zhmoginov, L.-C. Chen, Mobilenetv2: Inverted
  residuals and linear bottlenecks, in: Proceedings of the IEEE conference on
  computer vision and pattern recognition, 2018, pp. 4510--4520.

\bibitem{huang18-ddsss}
Z.~Huang, N.~Wang, Data-driven sparse structure selection for deep neural
  networks, in: ECCV, 2018, pp. 1--10.

\bibitem{pintor2021fast}
M.~Pintor, F.~Roli, W.~Brendel, B.~Biggio, Fast minimum-norm adversarial
  attacks through adaptive norm constraints, in: NeurIPS, 2021, pp. 1--10.

\bibitem{carlini19_onevaluating}
N.~Carlini, A.~Athalye, N.~Papernot, W.~Brendel, J.~Rauber, D.~Tsipras, I.~J.
  Goodfellow, A.~Madry, A.~Kurakin, On evaluating adversarial robustness, CoRR
  abs/1902.06705 (2019) 1--10.
\newblock \href {http://arxiv.org/abs/1902.06705} {\path{arXiv:1902.06705}}.

\bibitem{mura24_hofmn}
R.~Mura, G.~Floris, L.~Scionis, G.~Piras, M.~Pintor, A.~Demontis, G.~Giacinto,
  B.~Biggio, F.~Roli, Ho-fmn: Hyperparameter optimization for fast minimum-norm
  attacks, Neurocomputing (2024) 128918.

\bibitem{blalock20_state}
D.~W. Blalock, J.~J.~G. Ortiz, J.~Frankle, J.~V. Guttag, What is the state of
  neural network pruning?, in: Proceedings of Machine Learning and Systems
  2020, MLSys, 2020, pp. 1--10.

\bibitem{yeom_21pruning}
S.-K. Yeom, P.~Seegerer, S.~Lapuschkin, A.~Binder, S.~Wiedemann, K.-R.
  M{\"u}ller, W.~Samek, Pruning by explaining: A novel criterion for deep
  neural network pruning, Pattern Recognition 115 (2021) 107899.

\bibitem{wang_adversarial_2018}
L.~Wang, G.~W. Ding, R.~Huang, Y.~Cao, Y.~C. Lui, {ADVERSARIAL} {ROBUSTNESS}
  {OF} {PRUNED} {NEURAL} {NETWORKS} (2018) 1--10.

\bibitem{guo_sparse_2019}
Y.~Guo, C.~Zhang, C.~Zhang, Y.~Chen, Sparse dnns with improved adversarial
  robustness, in: NeurIPS, 2018, pp. 1--10.

\bibitem{wijayanto_robustness_2018}
A.~W. Wijayanto, J.~J. Choong, K.~Madhawa, T.~Murata, Robustness of compressed
  convolutional neural networks, in: International Conference on Big Data,
  2018, pp. 1--10.

\bibitem{sehwag19_towards}
V.~Sehwag, S.~Wang, P.~Mittal, S.~Jana, Towards compact and robust deep neural
  networks, CoRR abs/1906.06110.

\bibitem{carmon19-unlabeled}
Y.~Carmon, A.~Raghunathan, L.~Schmidt, P.~Liang, J.~Duchi, Unlabeled data
  improves adversarial robustness, in: NeurIPS, 2019, pp. 1--10.

\bibitem{zhao22_heracles}
Q.~Zhao, T.~K\"onigl, C.~Wressnegger, Non-uniform adversarially robust pruning,
  in: Proceedings of the First International Conference on Automated Machine
  Learning, PMLR, 2022, pp. 1--10.

\bibitem{wang20-mart}
Y.~Wang, D.~Zou, J.~Yi, J.~Bailey, X.~Ma, Q.~Gu, Improving adversarial
  robustness requires revisiting misclassified examples, in: ICLR, 2020, pp.
  1--10.

\bibitem{cui21_lbgat}
J.~Cui, S.~Liu, L.~Wang, J.~Jia, Learnable boundary guided adversarial
  training, in: ICCV, 2021, pp. 1--10.

\bibitem{lee_masking_2022}
B.-K. Lee, J.~Kim, Y.~M. Ro, Masking {Adversarial} {Damage}: {Finding}
  {Adversarial} {Saliency} for {Robust} and {Sparse} {Network}, in: CVPR, 2022,
  pp. 1--10.

\bibitem{wong20_fastat}
E.~Wong, L.~Rice, J.~Z. Kolter, Fast is better than free: Revisiting
  adversarial training, in: ICLR, 2020, pp. 1--10.

\bibitem{rakin19_robust}
A.~S. Rakin, Z.~He, L.~Yang, Y.~Wang, L.~Wang, D.~Fan, Robust sparse
  regularization: Simultaneously optimizing neural network robustness and
  compactness, CoRR abs/1905.13074.
\newblock \href {http://arxiv.org/abs/1905.13074} {\path{arXiv:1905.13074}}.

\bibitem{he19_cni}
Z.~He, A.~S. Rakin, D.~Fan, Parametric noise injection: Trainable randomness to
  improve deep neural network robustness against adversarial attack, in: CVPR,
  2019, pp. 1--10.

\bibitem{wei_batch_2021}
X.~Wei, Y.~Zhu, S.-T. Xia, Batch {Normalization} {Assisted} {Adversarial}
  {Pruning}: {Towards} {Lightweight}, {Sparse} and {Robust} {Models}, in: 2021
  {IJCNN}, Shenzhen, China, 2021, pp. 1--10.

\bibitem{lim_robustness-aware_2021}
H.~Lim, S.-D. Roh, S.~Park, K.-S. Chung, Robustness-{Aware} {Filter} {Pruning}
  for {Robust} {Neural} {Networks} {Against} {Adversarial} {Attacks}, in: 2021
  {IEEE} {MLSP}, 2021, pp. 1--10.

\bibitem{ilyas19_bugs}
A.~Ilyas, S.~Santurkar, D.~Tsipras, L.~Engstrom, B.~Tran, A.~Madry, Adversarial
  examples are not bugs, they are features, in: NeurIPS, 2019, pp. 1--10.

\bibitem{kaur_deadwooding_2022}
S.~Kaur, F.~Fioretto, A.~Salekin, Deadwooding: {Robust} {Global} {Pruning} for
  {Deep} {Neural} {Networks}, arXiv:2202.05226 [cs] (Sep. 2022).

\bibitem{goodfellow15-iclr}
I.~J. Goodfellow, J.~Shlens, C.~Szegedy, Explaining and harnessing adversarial
  examples, in: ICLR, 2015, pp. 1--10.

\bibitem{zhuang23_adversarial}
X.~Zhuang, Y.~Ge, B.~Zheng, Q.~Wang, Adversarial network pruning by filter
  robustness estimation, in: ICASSP, 2023, pp. 1--10.

\bibitem{luo23_towards}
H.~Luo, Z.~Zhuang, Y.~Li, M.~Tan, C.~Chen, J.~Zhang, Towards compact and robust
  model learning under dynamically perturbed environments, IEEE Transactions on
  Circuits and Systems for Video Technology (2023) 1--1\href
  {http://dx.doi.org/10.1109/TCSVT.2023.3337538}
  {\path{doi:10.1109/TCSVT.2023.3337538}}.

\bibitem{zhong2023adv_robust_gkp}
S.~Zhong, Z.~You, J.~Zhang, S.~Zhao, Z.~LeClaire, Z.~Liu, D.~Zha, V.~Chaudhary,
  S.~Xu, X.~Hu, One less reason for filter pruning: Gaining free adversarial
  robustness with structured grouped kernel pruning, in: NeurIPS, 2023.

\bibitem{qian_robust_2023}
Y.~Qian, W.~Huang, T.~Yao, K.~Chen, X.~Ling, B.~Wang, Z.~Gu, J.~Zhang, Robust
  {Filter} {Pruning} {Guided} by {Deep} {Frequency}-{Features} for {Edge}
  {Intelligence} (2023).

\bibitem{li_towards_2020}
B.~Li, S.~Wang, Y.~Jia, Y.~Lu, Z.~Zhong, L.~Carin, S.~Jana, Towards {Practical}
  {Lottery} {Ticket} {Hypothesis} for {Adversarial} {Training},
  arXiv:2003.05733 (Mar. 2020).

\bibitem{wang_achieving_2020}
S.~Wang, N.~Liao, L.~Xiang, N.~Ye, Q.~Zhang, Achieving {Adversarial}
  {Robustness} via {Sparsity}, Machine Learning 111~(2), arXiv:2009.05423 [cs,
  stat].

\bibitem{fu21_rst}
Y.~Fu, Q.~Yu, Y.~Zhang, S.~Wu, X.~Ouyang, D.~D. Cox, Y.~Lin, Drawing robust
  scratch tickets: Subnetworks with inborn robustness are found within randomly
  initialized networks, in: NeurIPS, 2021, pp. 1--10.

\bibitem{chen_sparsity_2021}
T.~Chen, Z.~Zhang, P.~Wang, S.~Balachandra, H.~Ma, Z.~Wang, Z.~Wang, Sparsity
  winning twice: Better robust generalization from more efficient training, in:
  ICLR, 2022, pp. 1--10.

\bibitem{shi22_finding}
X.~Shi, P.~Zheng, A.~A. Ding, Y.~Gao, W.~Zhang, Finding dynamics preserving
  adversarial winning tickets, in: {AISTATS}, 2022, pp. 1--10.

\bibitem{li_twinrep_2023}
C.~Li, Q.~Qiu, Z.~Zhang, J.~Guo, X.~Cheng, Learning {Adversarially} {Robust}
  {Sparse} {Networks} via {Weight} {Reparameterization}, Proceedings of the
  AAAI Conference on Artificial Intelligence.

\bibitem{ozdenizci_training_2021}
O.~Özdenizci, R.~Legenstein, Training {Adversarially} {Robust} {Sparse}
  {Networks} via {Bayesian} {Connectivity} {Sampling}, in: ICML, 2021, pp.
  1--10.

\bibitem{kurakin17_iclr}
A.~Kurakin, I.~J. Goodfellow, S.~Bengio, Adversarial machine learning at scale,
  in: ICLR, 2017.

\bibitem{kundu_tunable_2020}
S.~Kundu, M.~Nazemi, P.~A. Beerel, M.~Pedram, Dnr: A tunable robust pruning
  framework through dynamic network rewiring of dnns, in: Proceedings of the
  26th Asia and South Pacific Design Automation Conference, 2021, pp. 344--350.

\bibitem{dettmers19_scratch}
T.~Dettmers, L.~Zettlemoyer, Sparse networks from scratch: Faster training
  without losing performance, CoRR abs/1907.04840.
\newblock \href {http://arxiv.org/abs/1907.04840} {\path{arXiv:1907.04840}}.

\bibitem{haocvpr16-filters}
H.~Li, A.~Kadav, I.~Durdanovic, H.~Samet, H.~P. Graf,
  \href{https://openreview.net/forum?id=rJqFGTslg}{Pruning filters for
  efficient convnets}, in: International Conference on Learning
  Representations, 2017, pp. 1--10.
\newline\urlprefix\url{https://openreview.net/forum?id=rJqFGTslg}

\bibitem{liu2023lessons}
S.~Liu, Z.~Wang, Ten lessons we have learned in the new "sparseland": A short
  handbook for sparse neural network researchers (2023).
\newblock \href {http://arxiv.org/abs/2302.02596} {\path{arXiv:2302.02596}}.

\bibitem{pintor22_indicators}
M.~Pintor, L.~Demetrio, A.~Sotgiu, A.~Demontis, N.~Carlini, B.~Biggio, F.~Roli,
  Indicators of attack failure: Debugging and improving optimization of
  adversarial examples, in: NeurIPS, 2022, pp. 1--10.

\bibitem{cina24attackbench}
A.~E. Cin{\`a}, J.~Rony, M.~Pintor, L.~Demetrio, A.~Demontis, B.~Biggio, I.~B.
  Ayed, F.~Roli, Attackbench: Evaluating gradient-based attacks for adversarial
  examples, arXiv preprint arXiv:2404.19460.

\bibitem{lee21_iclr}
J.~Lee, S.~Park, S.~Mo, S.~Ahn, J.~Shin, Layer-adaptive sparsity for the
  magnitude-based pruning, in: International Conference on Learning
  Representations, 2021, pp. 1--10.

\bibitem{jordao_effect_2021}
A.~Jord{\~{a}}o, H.~Pedrini, On the effect of pruning on adversarial
  robustness, in: ICCVW, 2021, pp. 1--10.

\bibitem{vora_benchmarking_2023}
B.~Vora, K.~Patwari, S.~M. Hafiz, Z.~Shafiq, C.-N. Chuah, Benchmarking
  {Adversarial} {Robustness} of {Compressed} {Deep} {Learning} {Models},
  arXiv:2308.08160 [cs] (Aug. 2023).

\bibitem{merkle_pruning_2021}
F.~Merkle, M.~Samsinger, P.~Sch{\"o}ttle, Pruning in the face of adversaries,
  in: ICIAP 2022, Springer International Publishing, Cham, 2022, pp. 1--10.

\bibitem{piras23_thinice}
G.~Piras, M.~Pintor, A.~Demontis, B.~Biggio, Samples on thin ice: Re-evaluating
  adversarial pruning of neural networks, in: {ICMLC}, {IEEE}, 2023, pp. 1--10.

\bibitem{pavlitska23_relationship}
S.~Pavlitska, H.~Grolig, J.~M. Z{\"{o}}llner, Relationship between model
  compression and adversarial robustness: {A} review of current evidence,
  CoRR\href {http://arxiv.org/abs/2311.15782} {\path{arXiv:2311.15782}}.

\bibitem{cheng2023survey}
H.~Cheng, M.~Zhang, J.~Q. Shi, A survey on deep neural network pruning:
  Taxonomy, comparison, analysis, and recommendations, IEEE Transactions on
  Pattern Analysis and Machine Intelligence.

\bibitem{chen21_chasing}
T.~Chen, Y.~Cheng, Z.~Gan, L.~Yuan, L.~Zhang, Z.~Wang, Chasing sparsity in
  vision transformers: An end-to-end exploration, NeurIPS 2021.

\end{thebibliography}

\end{document}